\documentclass[runningheads]{llncs}

\usepackage{eccv}
\usepackage{eccvabbrv}
\usepackage{graphicx}
\usepackage{booktabs}
\usepackage[accsupp]{axessibility}

\usepackage{algorithm}
\usepackage{algpseudocode}
\algnewcommand\algorithmicinput{\textbf{Input:}}
\algnewcommand\Input{\item[\algorithmicinput]}
\algnewcommand\algorithmicoutput{\textbf{Output:}}
\algnewcommand\Output{\item[\algorithmicoutput]}
\usepackage{float}
\usepackage{amsmath}
\usepackage{amssymb}
\makeatletter
\newenvironment{breakablealgorithm}{%
  \begin{center}
    \refstepcounter{algorithm}%
    \hrule height.8pt depth0pt \kern2pt
    \renewcommand{\caption}[2][\relax]{%
      {\raggedright\textbf{\ALG@name~\thealgorithm} ##2\par}%
      \ifx\relax##1\relax
        \addcontentsline{loa}{algorithm}{\protect\numberline{\thealgorithm}##2}%
      \else
        \addcontentsline{loa}{algorithm}{\protect\numberline{\thealgorithm}##1}%
      \fi
      \kern2pt\hrule\kern2pt
    }
}{%
    \kern2pt\hrule\relax
  \end{center}
}
\makeatother

\usepackage{hyperref}
\usepackage{orcidlink}
\usepackage{amsmath}
\usepackage{amssymb}
\usepackage{multirow}
\usepackage{colortbl}
\usepackage{marvosym}
\renewcommand\footnotemark{}

\begin{document}

\title{Anchoring and Steering Diffusion: \\ Enhancing the Faithfulness of \\ Text-to-Image Generation at Inference Time}

\titlerunning{Anchoring and Steering Diffusion}
\author{
Xinyi Wang \and 
Yuyang Huang \and
Yalin Su\and 
Pengcheng Luan\and \\ 
Tao Zhang\textsuperscript{(\Letter)} \and 
Feiming Wei \and 
Wenxian Yu
\thanks{Corresponding author: Tao Zhang.}
}

\authorrunning{X. Wang et al.}

\institute{Shanghai Jiao Tong University, Shanghai, China \\
\email{\{xinyi\_wang, huangyuyang, syl\_move, pengcheng\_luan,
\\sjtu-{}-zt, weifeiming, wxyu\}@sjtu.edu.cn}\\
}

\maketitle

\begin{figure}[!ht]
    \centering
    \includegraphics[width=\textwidth]{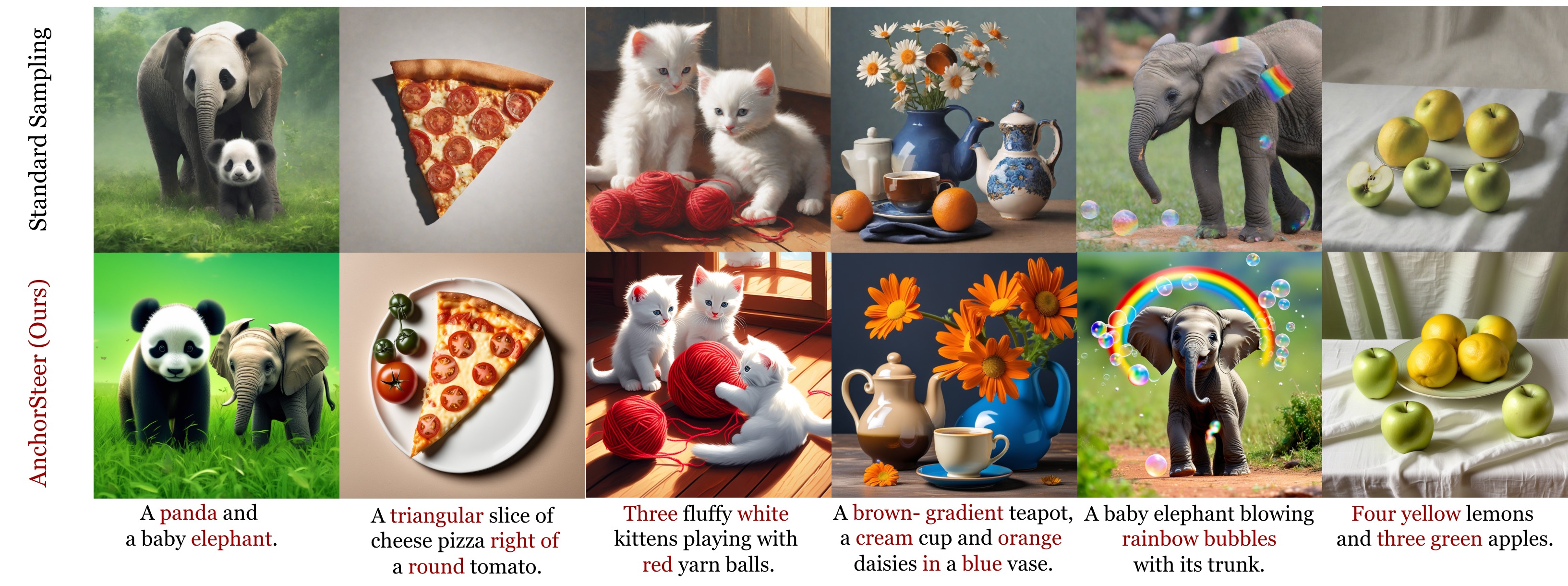}
\caption{Visual comparison of text-to-image results. Our \textbf{AnchorSteer} demonstrates better text–image alignment across various aspects.
}
    \label{fig:teaser}
\end{figure}

\begin{abstract}
While text-to-image diffusion models achieve impressive visual quality, they frequently struggle to maintain precise alignment with complex compositional prompts. An effective strategy is to improve the inference process of diffusion models, thereby better leveraging their pretrained priors to address misalignment. Existing training-free methods can be divided into two categories. The first category focuses on improving the randomly sampled initial noise to obtain an initialization that encapsulates semantics relevant to the target text. However, existing approaches either perform costly search over noise pools, with no guarantee of finding a truly prompt-compatible noise within the limited candidates, or manipulate sampled noise without simultaneously ensuring reliable semantic injection and preservation of the Gaussian distribution. The second category focuses on improving the denoising trajectory. However, these methods lack explicit mechanisms to timely diagnose and correct semantic errors, and therefore fail to prevent error propagation during generation. To address these limitations, we propose \textbf{AnchorSteer}, a training-free framework that exerts fine-grained control over \textbf{both initialization} and \textbf{the denoising trajectory}. AnchorSteer consists of two synergistic components:
\textbf{Semantic Anchoring} replaces uninformative Gaussian noise with text-aligned initializations via CLIP-based prior extraction and a novel Latent-Prior Score Distillation Sampling (LP-SDS) objective. 
Specifically, LP-SDS distills CLIP visual priors into the knowledge distribution of diffusion models, mitigating the domain gap between CLIP-based priors and diffusion-based priors.
\textbf{Reflective Steering} transforms passive denoising with an active \textbf{Think--Erase--Retouch loop} that enables mid-generation self-correction. \textit{Think} phase employs VLM-based dual diagnosis to detect semantic deviations. \textit{Erase} phase performs a targeted latent rollback with negative-guided inversion to suppress erroneous content, and \textit{Retouch} phase subsequently applies positive-guided refinement to recover the missing attributes. Therefore, \textbf{Reflective Steering} enables timely correction of semantic deviations along the denoising trajectory, preventing error propagation. 
Extensive experiments on GenEval and T2I-CompBench++ demonstrate that AnchorSteer consistently outperforms existing baselines in text--image alignment while preserving high visual quality.
  
\end{abstract}

\section{Introduction}
\label{sec:intro}
Text-to-image (T2I) diffusion models~\cite{ho2020denoising, rombach2022high, saharia2022photorealistic,song2021scorebased} have achieved remarkable success in generating high-quality images. However, they still frequently encounter challenges in maintaining precise semantic alignment~\cite{chefer2023attend,feng2023trainingfree,wang2024compositional,ghosh2023geneval,huang2025t2icompbench++,rassin2022dalle}, particularly \textbf{under complex compositional prompts} ( e.g., involving multiple objects, attributes, and spatial constraints). As is shown in Figure~\ref{fig:teaser}, under such scenarios, key semantic details are often lost or distorted, leading to missing objects, attribute leakage, or spatial inconsistencies~\cite{liu2024correcting,chang2024repairing,rassin2023linguistic,zhuang2024magnet,jiang2024comat,chatterjee2024getting}. 

To address these issues, existing works can be broadly categorized into two groups: \textit{training-based} and \textit{training-free} methods. Training-based methods improve text–image alignment by fine-tuning diffusion models on customized image–text datasets involving complex semantics~\cite{ruiz2023dreambooth,brooks2023instructpix2pix,kumari2023multi}. 
However, the diversity of possible compositional scenarios is effectively unbounded, making it difficult for the training distribution to cover all cases. 
As a result, these approaches often exhibit limited generalization to unseen combinations, which restricts their overall effectiveness.

Recent studies observe that pretrained diffusion models already demonstrate strong capabilities in handling individual sub-prompts~\cite{lian2024llmgrounded,liu2022compositional,chefer2023attend,hertz2023prompttoprompt}, suggesting that improving the inference process can serve as an effective complement to training-based methods for enhancing alignment with complex textual prompts. This insight motivates training-free methods that improve text–image alignment by enhancing the inference process and better utilizing pretrained models. 

\begin{figure}[t]
    \centering
    \includegraphics[
    width=\linewidth,
    trim=0cm 0cm 0cm 0cm,
    clip
]{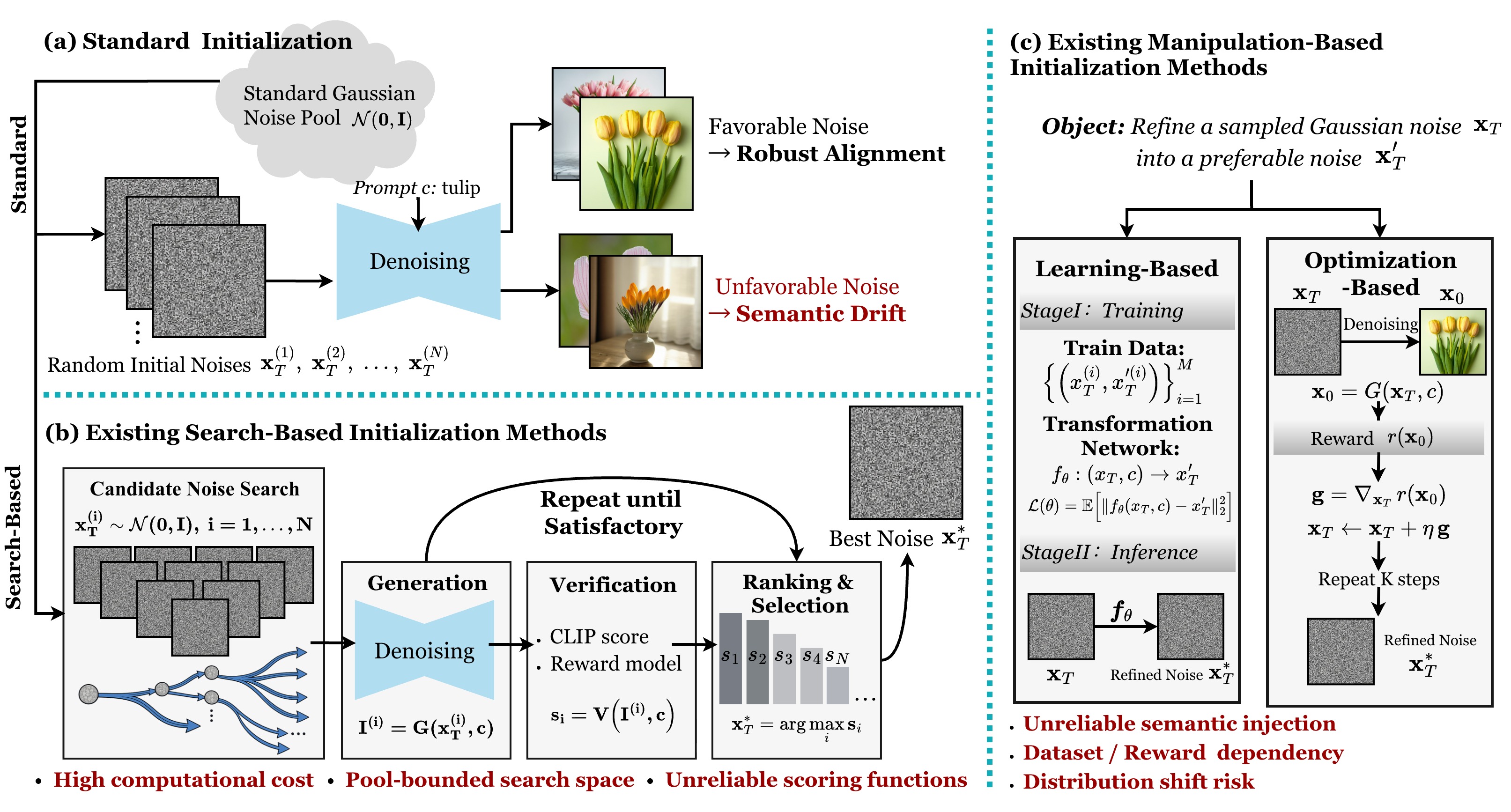}
\caption{
\textbf{Existing methods for noise initialization and their  limitations.}
(a) Standard Gaussian initialization treats all noises as semantically equivalent, potentially assigning an unfavorable noise to the prompt and causing semantic drift during denoising.
(b) Search-based methods rely on candidate selection from a predefined noise pool, leading to high computational cost and unreliable alignment scoring.
(c) Manipulation-based approaches refine sampled noise via learning or optimization, yet suffer from unstable semantic injection and potential distribution shift.
}
    \label{fig:init_comparison}
\end{figure}

To begin with, misalignment during inference mainly stems from two components. 
First, at \textbf{\textit{noise initialization}}, randomly sampling from standard Gaussian sampling often produces starting points unfavorable for complex compositional prompts(Figure~\ref{fig:init_comparison}a). 
Second, along the \textbf{\textit{denoising trajectory}}, the inference process lacks mechanisms to detect or correct semantic deviations, hindering the satisfaction of multiple prompt constraints. Therefore, existing training-free methods mainly focus on improving these two components.

For \textit{noise initialization}, recent studies\cite{liu2024correcting,samuel2024generating,kim2026model, ma2025inference, ahn2026a,zhou2025golden,
zeng2025d2dpm,tang2025inferencetime,harrington2026s} 
has focused on constructing initial noise that encapsulates semantics aligned with the given text prompt, thereby improving the generated results. Existing methods can be broadly grouped into two categories.

The first category consists of \textbf{search-based methods}(Figure~\ref{fig:init_comparison}b)~\cite{liu2024correcting,samuel2024generating,kim2026model, ma2025inference}, which aim to identify a prompt-compatible initial noise by selecting the best candidate from a predefined noise pool. However, search-based methods are inefficient and fundamentally limited by the noise pool and imperfect alignment scoring employed for searching. Another category is \textbf{manipulation-based methods} (Figure~\ref{fig:init_comparison}c), which refine a randomly sampled initial noise into a more prompt-compatible one~\cite{ahn2026a,zhou2025golden,zeng2025d2dpm,tang2025inferencetime,harrington2026s}. However, manipulation-based methods cannot simultaneously guarantee reliable semantic injection and preservation of the Gaussian noise distribution after manipulation.

For \textit{denoising trajectory}, existing methods can be broadly categorized into two groups: unidirectional and bidirectional. 
\textbf{Unidirectional methods} attempt to optimize the trajectory within a single forward pass through various strategies, such as increasing denoising steps~\cite{grimal2025texttoimagealignmentdenoisingbasedmodels} or scaling up classifier-free guidance~\cite{chung2025cfg,shen2024rethinking,sadat2024eliminating}. 
However, they still cannot recover from intermediate semantic deviations. 
\textbf{Bidirectional methods}~\cite{song2021denoising,karras2022elucidating,feng2024explorative,bai2024zigzag} introduce re-noising and re-denoising transitions, allowing them to revisit earlier states. 
However, they still lack explicit mechanisms to diagnose and correct semantic errors. 
Therefore, semantic errors may propagate and accumulate along the denoising trajectory.

To tackle the aforementioned limitations in \textit{noise initialization} and \textit{denoising trajectory}, we propose \textbf{AnchorSteer}, a training-free framework that exerts fine-grained control over both noise initialization and denoising trajectory. To replace unreliable noise initialization, we introduce \textbf{Semantic Anchoring}, which constructs initial noise that is both text-aligned and distributionally consistent. To overcome the lack of reliable error rectification during the denoising trajectory, we further develop \textbf{Reflective Steering}, a strategy that enables faithful diagnosis and targeted correction of semantic deviations throughout the diffusion process. 
Extensive experiments show that AnchorSteer achieves superior alignment across different diffusion models without training.

The main contributions in our paper are summarized as follows:

\begin{itemize}
    \item We propose \textbf{AnchorSteer}, a training-free framework that explicitly enhances initial noise with text-aligned semantics and performs timely rectification of the denoising trajectory to prevent the propagation of semantic deviations.

\item We propose \textbf{Semantic Anchoring}, a principled initialization strategy that first extracts a deterministic semantic prior via CLIP, then bridges the domain gap between CLIP outputs and the diffusion latent manifold through \textbf{Latent-Prior Score Distillation Sampling (LP-SDS)}, and finally applies DDPM forward diffusion to yield initial noise that is both text-aligned and distributionally consistent.

\item We introduce \textbf{Reflective Steering}, a bidirectional self-corrective mechanism that diagnoses and corrects semantic deviations via a \textit{Think--Erase--Retouch} loop, enabling explicit suppression of inconsistent content and reinforcement of missing semantics.
\end{itemize}

\section{Related Work}
\label{sec:related}

\noindent
\textbf{Text-to-Image Diffusion Models.}
Image diffusion models have achieved remarkable success in text-to-image generation~\cite{ho2020denoising,song2021scorebased} and a wide range of other visual tasks~\cite{tian2024diffuse,ke2024repurposing,huang2024domainfusion}. Latent Diffusion Models~\cite{rombach2022high} such as SDXL~\cite{podell2024sdxl} improve efficiency via latent-space generation, while recent Transformer-based architectures~\cite{peebles2023scalable,esser2024scaling,flux2024} further enhance scalability and visual quality. Despite these advances, models still struggle with complex compositional prompts, often producing missing objects, attribute leakage, or spatial inconsistencies~\cite{liu2024correcting,rassin2023linguistic,jiang2024comat,zhang2025compass}.

\noindent
\textbf{Noise Initialization Strategies.}
Recent studies~\cite{qi2024noisescreatedequallydiffusionnoise,guo2024initno} show that different initial noises lead to different generation outcomes. 
\textit{Search-based methods}~\cite{liu2024correcting,samuel2024generating,ma2025inference} select favorable samples from noise pools but are computationally expensive and limited by pool diversity. 
\textit{Manipulation-based methods} modify noise via learned mappings~\cite{ahn2026a,zhou2025golden} or gradient optimization~\cite{zeng2025d2dpm,tang2025inferencetime}, yet often lack reliable semantic injection or introduce distribution shift. 
Our LP-SDS addresses both issues by ensuring semantic alignment while remaining consistent with the diffusion model's training distribution.

\noindent
\textbf{Inference-Time Trajectory Optimization.}
Prior work improves alignment by refining the denoising trajectory at inference time. \textit{Unidirectional methods} adjust guidance strength or introduce gradient-based updates~\cite{chung2025cfg,sadat2024eliminating,yu2023freedom}, but cannot recover once the trajectory deviates from the desired semantics. \textit{Bidirectional methods}~\cite{bai2024zigzag,huang2025zero,huang2026diffusion} introduce re-noising steps to revisit earlier states, yet lack explicit mechanisms for diagnosing and correcting semantic errors. In contrast, our Reflective Steering introduces a VLM-guided Think–Erase–Retouch loop that actively detects and rectifies semantic inconsistencies.

\section{Preliminaries}
\label{sec:prelim}

\subsection{Latent Diffusion Models}
\label{sec:prelim_diffusion}

Latent Diffusion Models (LDMs)~\cite{rombach2022high} operate in the
latent space of a pretrained autoencoder
$(\mathcal{E},\,\mathcal{D})$.  A clean latent
$z_0\!=\!\mathcal{E}(x)$ is corrupted by the forward process $z_t = \sqrt{\bar\alpha_t}\,z_0
       +\sqrt{1-\bar\alpha_t}\,\epsilon,
  \quad \epsilon\sim\mathcal{N}(0,\mathbf{I})$,
where $\bar\alpha_t=\prod_{s=1}^{t}\alpha_s$. A text-conditional
denoising network $\epsilon_\phi$ is trained to predict the added
noise by minimizing~\cite{song2021scorebased}
\begin{equation}
  \mathcal{L}_{\mathrm{Diff}}
  =\mathbb{E}_{t,\epsilon}\!\bigl[
      w(t)\,\bigl\|
          \epsilon_\phi(z_t;t,c)-\epsilon
      \bigr\|_2^2
  \bigr],
  \label{eq:diff_objective}
\end{equation}
with text condition $c$ and time-dependent weight $w(t)$.
Starting from $z_T\!\sim\!\mathcal{N}(0,\mathbf{I})$, the reverse
process iteratively denoises with $\epsilon_\phi$.
Classifier-Free Guidance (CFG)~\cite{ho2021classifierfree} enhances text
alignment by extrapolating between conditional and unconditional
estimates with guidance scale $s$:
\begin{equation}
  \hat\epsilon_\phi(z_t;t,c)
  =\epsilon_\phi(z_t;t,\varnothing)
   +s\bigl(
       \epsilon_\phi(z_t;t,c)
      -\epsilon_\phi(z_t;t,\varnothing)
   \bigr),\quad s>1.
  \label{eq:cfg}
\end{equation}

\subsection{Score Distillation Sampling}
\label{sec:prelim_sds}

Score Distillation Sampling
(SDS)~\cite{poole2023dreamfusion} optimizes a differentiable
representation $x\!=\!g(\theta)$ using a frozen diffusion model as a
probabilistic critic.  After encoding and perturbing
$z_0\!=\!\mathcal{E}(x)$ to $z_t$,
the parameter gradient is
\begin{equation}
  \nabla_\theta\mathcal{L}_{\mathrm{SDS}}
  =\mathbb{E}_{t,\epsilon}\!\Bigl[
       w(t)\bigl(
          \hat\epsilon_\phi(z_t;t,c)-\epsilon
      \bigr)
      \tfrac{\partial x}{\partial\theta}
  \Bigr],
  \label{eq:sds}
\end{equation}
which drives $x$ toward the natural-image manifold consistent
with~$c$ by distilling the generative prior of the frozen model, where constant scaling factors are absorbed into $w(t)$.

\section{Methodology}
\begin{figure}[t]
    \centering
    \includegraphics[width=\textwidth]{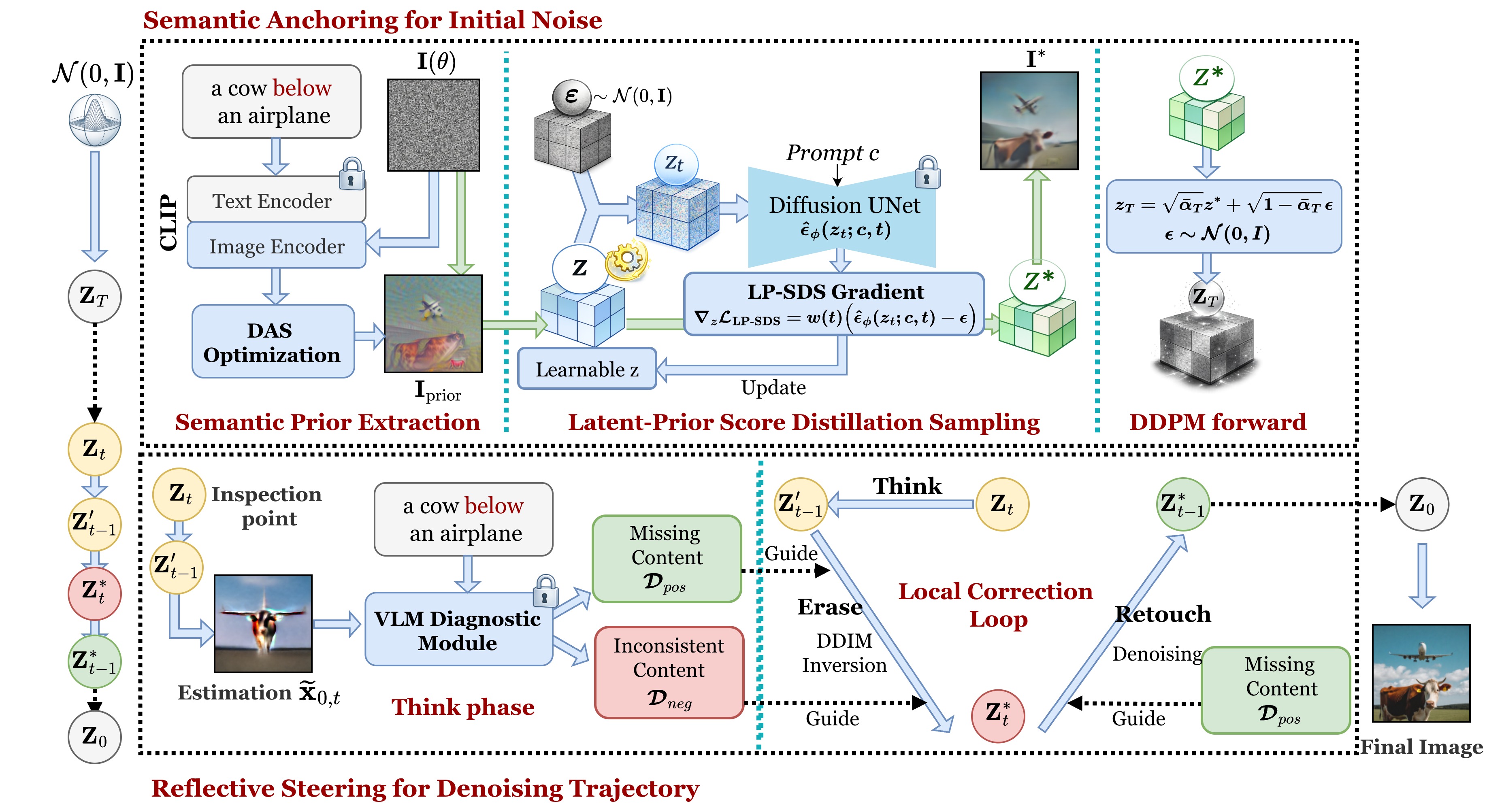}
    \caption{
    \textbf{Overview of AnchorSteer.}
    \textbf{(Top) Semantic Anchoring for Initial Noise
    (Section~\ref{sec:method_1}).}
    Given a text prompt $c$, a semantic prior $I_{\mathrm{prior}}$ is
    first extracted from the CLIP model via Direct Ascent Synthesis
    (DAS), then encoded into the latent space and refined through Latent-Prior Score Distillation Sampling (LP-SDS) to obtain an optimized latent
    $z^*$ that bridges the domain gap. The standard DDPM forward
    noising process is finally applied to produce an initial noise
    $z_T$ that is both semantically aligned and distributionally
    consistent.
    \textbf{(Bottom) Reflective Steering for Denoising Trajectory
    (Section~\ref{sec:method_2}).}
    At selected timesteps, a Think--Erase--Retouch loop is inserted
    into the denoising trajectory $z_t \!\rightarrow\! z_{t-1}$.
    The \textit{Think} module employs a VLM to diagnose the current
    estimate $z_{t-1}'$, producing a negative prompt
    $\mathcal{D}_{\mathrm{neg}}$ (inconsistent content) and a positive
    prompt $\mathcal{D}_{\mathrm{pos}}$ (missing content).
    The \textit{Erase} module performs negative-guided DDIM inversion
    ($z_{t-1}' \!\rightarrow\! z_{t}^{*}$) to suppress erroneous
    semantics.
    The \textit{Retouch} module applies positive guidance during
    re-noising and re-denoising
    ($z_{t-1}' \!\rightarrow\! z_{t}^{*} \!\rightarrow\! z_{t-1}^{*}$)
    to reinforce missing content, actively correcting semantic drift.
    }
    \label{fig:methods}
\end{figure}

Our framework \textbf{AnchorSteer} \textbf{anchors the initial noise} for
\textbf{a good start} and reflectively \textbf{steers the denoising
trajectory} for \textbf{precise guidance}. The overall framework is
illustrated in Figure~\ref{fig:methods}.

\subsection{Semantic Anchoring for Initial Noise} \label{sec:method_1}

To ensure effective semantic injection, we \textbf{extract text-aligned semantic prior} from \textbf{the CLIP}~\cite{radford2021learning}, thereby providing a deterministic and reliable semantic enhancement to the manipulated noise. To eliminate the domain gap of the noise distribution, we further propose \textbf{Latent-Prior Score Distillation Sampling (LP-SDS)} to refine
this prior and apply the DDPM forward process to inject the refined
semantic prior into the initial noise.

\subsubsection{\textbf{CLIP-based Semantic Prior Extraction.}}

\begin{figure}[t]
\centering
\includegraphics[width=\linewidth]{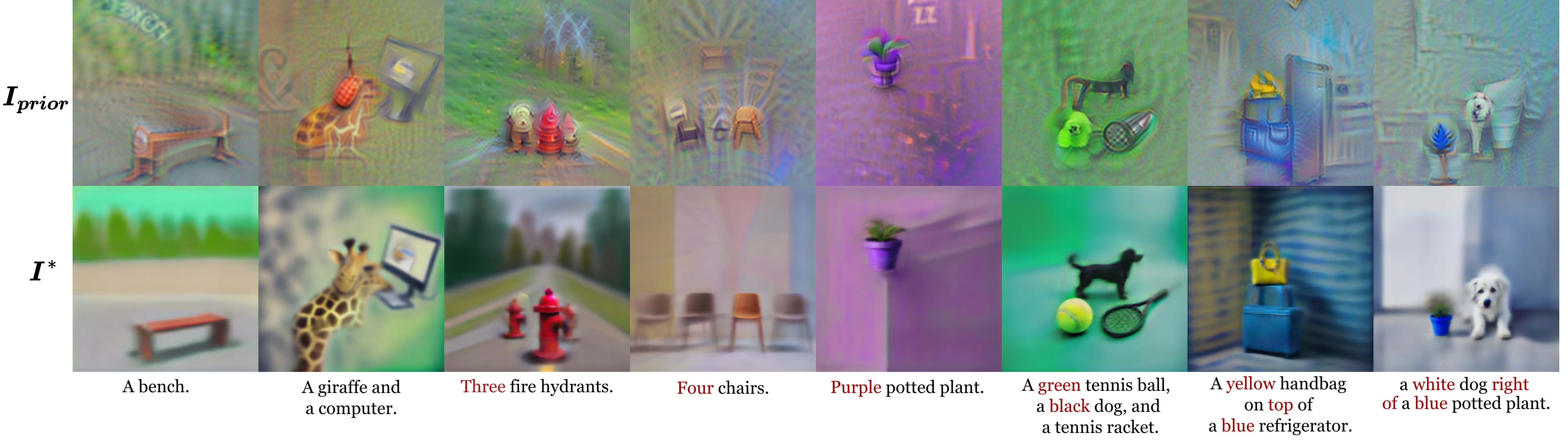}
\caption{
\textbf{CLIP-based semantic priors and their refinement via LP-SDS.} The first row shows CLIP semantic priors $I_{\mathrm{prior}}$, which are semantically aligned but OOD for diffusion models.
The second row shows refined images $I^*$ after LP-SDS, which become in-distribution while preserving the same semantics.
}
\label{fig:clip_prior}
\end{figure}

CLIP encodes rich semantic knowledge about objects, attributes, and relational constraints for nearly arbitrary prompts. We therefore seek to extract these semantics for complex prompts and leverage them to construct the initial noise. Inspired by Direct Ascent Synthesis
(DAS)~\cite{fort2025directascentsynthesisrevealing} to extract the
semantic prior embedded within CLIP. Given a target text prompt
$c$, we parameterize a learnable image $I(\theta)$ and directly
optimize it to maximize its alignment with $c$ in the CLIP
embedding space using its image encoder $\mathcal{E}_I(\cdot)$ and text
encoders $\mathcal{E}_T(\cdot)$ by $
    \mathcal{L}_{\mathrm{DAS}}(\theta)
    = -[
        \mathcal{E}_I(I(\theta))
        \cdot
        \mathcal{E}_T(c)
      ]/[
        \bigl\|\mathcal{E}_I(I(\theta))\bigr\|
        \;\bigl\|\mathcal{E}_T(c)\bigr\|
      ].$
After $N_{\mathrm{das}}$ steps, the resulting pseudo-image
$I_{\mathrm{prior}} = I(\theta_{N_{\mathrm{das}}})$ captures spatial
structure and color distribution that are deterministically anchored
to the semantics of $c$.

\subsubsection{\textbf{Bridging Domain Gap via LP-SDS.}}
\begin{figure}[t]
    \centering
    \includegraphics[
        width=\linewidth,
        trim=0cm 0cm 0cm 0cm,
        clip
    ]{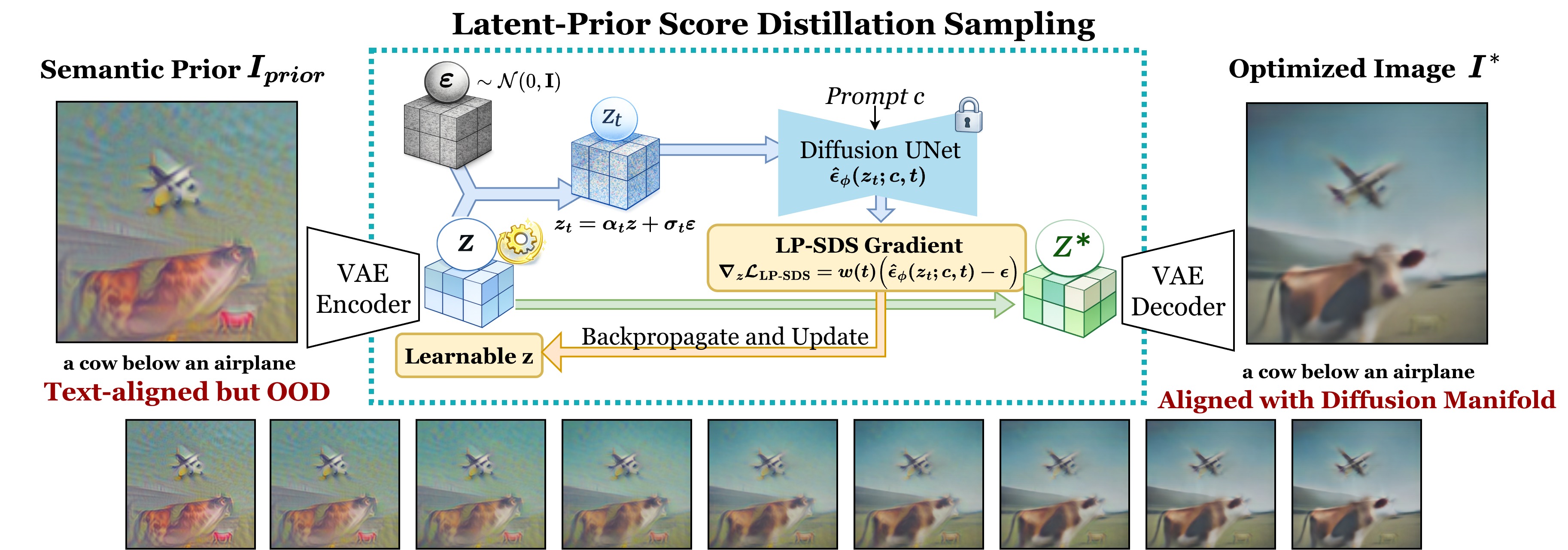}
    \caption{
    \textbf{Illustration of LP-SDS.}
    The semantic prior $I_{\mathrm{prior}}$ produced by DAS is encoded
    into the latent space as $z_0 = \mathcal{E}_{\text{VAE}}(I_{\mathrm{prior}})$
    and iteratively refined via LP-SDS gradients from a frozen diffusion U-Net.
    The optimized latent $z^*$ is projected onto the diffusion manifold,
    bridging the domain gap between CLIP-based synthesis and
    diffusion-based generation while preserving text-aligned semantics.
    }
    \label{fig:lp}
\end{figure}

As illustrated in Fig.~\ref{fig:clip_prior}, the CLIP-based pseudo-images 
$I_{\mathrm{prior}}$ exhibits severe visual degradation, including high-frequency artifacts and unnatural color distributions, despite preserving text-aligned semantic structures. 
Such images deviate significantly from the natural image distribution learned by pre-trained diffusion models.
To mitigate the domain gap, we propose \textbf{Latent-Prior Score Distillation Sampling (LP-SDS)}, illustrated in Fig.~\ref{fig:lp},  which extends Score Distillation Sampling (SDS)~\cite{poole2023dreamfusion} to the noise optimization setting. 

We initialize the optimization variable as $z_0 = \mathcal{E}_{\mathrm{VAE}}(I_{\mathrm{prior}})$. We add noise to $z_0$ and input it to the diffusion model to reconstruct $z_0$. Then we compute the gradient of the diffusion loss $\mathcal{L}_{\mathrm{Diff}}$ with respect to $z$. 
Following SDS~\cite{poole2023dreamfusion}, we omit the U-Net Jacobian term, then the gradient simplifies to:
\begin{equation}
\nabla_{z}\mathcal{L}_{\mathrm{LP\text{-}SDS}}(\phi, z)
\triangleq
\mathbb{E}_{t,\epsilon}\left[
{w}(t)\,
\left(\hat{\epsilon}_\phi(z_t;\,c,t)-\epsilon\right)
\right].
\label{eq:sds_grad}
\end{equation}
From a theoretical perspective, it can be proved that the gradient implicitly minimizes a Kullback--Leibler (KL) divergence weighted across time steps~\cite{poole2023dreamfusion}:
\begin{equation}
\nabla_{z}\mathcal{L}_{\mathrm{LP\text{-}SDS}}(\phi, z)
=
\nabla_{z}\,
\mathbb{E}_{t}
\left[
\frac{\sigma_t}{\alpha_t}\, w(t)\,
\mathrm{KL}\!\left(
q(z_t\mid z)\,\|\,p_\phi(z_t;\,c,t)
\right)
\right],
\end{equation}
where $q(z_t\mid z)=\mathcal{N}(\alpha_t z,\,\sigma_t^2 I)$ is the Gaussian distribution induced by the forward diffusion process from the current latent variable $z$, and $p_\phi(z_t;\,c,t)$ is the marginal distribution conditioned on text $c$ at time step $t$. 
Thus, this optimization can be viewed as a \textbf{probability density distillation process }, driving $z$ towards high-density regions of the conditional distribution $p_\phi(z_t;\,c,t)$, thereby regularizing the optimized latent to remain consistent with the generative prior of the diffusion model.

After $ N_{\mathrm{lp}}$ steps of optimization, we obtain a refined image $I^*$. As shown in Fig.~\ref{fig:clip_prior}, $I^*$
exhibit significantly improved visual coherence compared with
the degraded priors $I_{\mathrm{prior}}$, while still preserving
the underlying text-aligned semantic structures.

\subsubsection{\textbf{DDPM Forward Diffusion for Noise Initialization.}}
We diffuse $z^*$ to the terminal step $T$ via the DDPM forward process to obtain the ultimate initial noise $z_T$ by $
  z_T = \sqrt{\bar{\alpha}_T}\, z^* + \sqrt{1 - \bar{\alpha}_T}\, \epsilon$. Therefore, $z_T$ still follows $\mathcal{N}(0, \mathbf{I})$.

\subsection{Reflective Steering  for Denoising Trajectory}\label{sec:method_2}

For denoising trajectory, we also adopt a fine-grained semantic correction method via a \textbf{Think–Erase–Retouch} loop. We evenly select inspection timesteps $\mathcal{T}_{\mathrm{insp}} \subset \{1, \dots, T\}$ with stride $\tau$. For each $t \in \mathcal{T}_{\mathrm{insp}}$, we expand the standard denoising trajectory into $z_{t} \rightarrow z_{t-1}' \rightarrow z_{t}^* \rightarrow z_{t-1}^*$, where $z_{t-1}'$ denotes the initial denoising estimate, and $z_{t}^*$, $z_{t-1}^*$ denote the refined latents after correction loop.

\subsubsection{\textbf{Think: Semantic Diagnosis via Vision–Language Feedback.}}

At each inspection timestep $t$, once the initial estimate ${z}_{t-1}'$ is obtained via standard denoising, the \textit{Think} phase performs an on-the-fly semantic diagnosis to assess its alignment with the text prompt. This involves two pivotal steps: Intermediate State Visualization and Dual Semantic Gap Diagnosis.

\textbf{1) Intermediate State Visualization.}
Following DDIM prediction, we first obtain the clean latent ${{z}}_{0|t-1}'$ of ${z}_{t-1}'$  and decode it into pixel space via the VAE decoder to yield the visualized image $\tilde{{x}}_{0|t-1}' = \mathcal{D}(\tilde{{z}}_{0|t-1}')$, which is then used for subsequent VLM-based diagnosis.

\textbf{2) Dual Semantic Gap Diagnosis.} Based on the estimated
$\tilde{{x}}_{0|t-1}'$, we leverage the robust cross-modal
understanding capabilities of a pre-trained Vision-Language Model (VLM)
to assess its alignment with the text prompt $c$. Departing from prior
methods that rely on implicit scalar scoring, we structure semantic
deviations into two distinct, actionable guidance signals:

\begin{itemize}
    \item \textbf{Missing Content ($\mathcal{D}_{\mathrm{pos}}$):}
    Identifies contents explicitly specified in $c$ but
    insufficiently manifested in $\tilde{{x}}_{0|t-1}'$
    (\emph{e.g.}, specific objects, attributes, or spatial
    relationships). $\mathcal{D}_{\mathrm{pos}}$ forms the target set
    for reinforcement in the subsequent \textit{Retouch} phase.

    \item \textbf{Inconsistent Content ($\mathcal{D}_{\mathrm{neg}}$):}
    Detects hallucinatory or extraneous content in
    $\tilde{{x}}_{0|t-1}'$ that contradicts or is irrelevant to
    $c$. $\mathcal{D}_{\mathrm{neg}}$ defines the target set for
    suppression in the subsequent \textit{Erase} phase.
\end{itemize}

Through this structured diagnosis, vague image-text misalignments are
transformed into precise, navigational correction signals that enable
targeted optimization of the denoising trajectory.

\subsubsection{\textbf{Erase and Retouch: Local Correction Loop for Targeted Semantic Refinement.}}

Upon acquiring the diagnostic sets $\mathcal{D}_{\mathrm{pos}}$ and
$\mathcal{D}_{\mathrm{neg}}$ from the Think stage, we perform a
local bidirectional correction at the inspection timestep $t$:
${z}_{t-1}' \xrightarrow{\text{Erase}} {z}_{t}^{*}
\xrightarrow{\text{Retouch}} {z}_{t-1}^{*}$. This loop
exploits the local reversibility of diffusion forward--reverse
processes, applying negative guidance during noising to suppress
errors and positive guidance during both noising and denoising to
reinforce missing attributes.

\textbf{1) Erase: Targeted Removal via Negative-Guided Inversion.}
This step executes the re-noising trajectory
$z_{t-1}' \rightarrow z_{t}^{*}$. We replace the unconditional
estimate $\epsilon_\phi(z_t;t,\varnothing)$ in Eq.~\eqref{eq:cfg}
with $\epsilon_\phi(z_t;t,\mathcal{D}_{\mathrm{neg}})$, steering
the latent away from the identified erroneous semantics. The guided
noise estimate $\hat{\boldsymbol{\epsilon}}_{\mathrm{e}}$ is computed
as:
\begin{equation}
\label{eq:erase_guidance}
\hat{\epsilon}_{\mathrm{e}}(z_{t-1}'; t{-}1)
\;=\;
(1+\gamma_{\mathrm{e}})\,
\epsilon_\phi(z_{t-1}'; t{-}1,\, c)
\;-\;
\gamma_{\mathrm{e}}\,
\epsilon_\phi(z_{t-1}'; t{-}1,\,
\mathcal{D}_{\mathrm{neg}})
\end{equation}
We then compute the predicted clean latent $\hat{z}_{0|t-1}'$ via Tweedie's formula and apply the DDIM inversion rule to obtain the re-noised state $z_{t}^{*}$:
\begin{equation}
\label{eq:erase_clean}
\hat{z}_{0|t-1}' \;=\;
\frac{z_{t-1}' - \sqrt{1-\bar{\alpha}_{t-1}}\;
\hat{\epsilon}_{\mathrm{e}}(z_{t-1}'; t{-}1)}
{\sqrt{\bar{\alpha}_{t-1}}}
\end{equation}
\begin{equation}
\label{eq:erase_step}
z_{t}^{*} \;=\;
\sqrt{\bar{\alpha}_{t}}\;\hat{z}_{0|t-1}'
\;+\;
\sqrt{1-\bar{\alpha}_{t}}\;
\hat{\epsilon}_{\mathrm{e}}(z_{t-1}'; t{-}1)
\end{equation}
By substituting $\mathcal{D}_{\mathrm{neg}}$ into the negative-prompt
position, the guidance gradient actively steers the inversion
trajectory away from the undesirable semantic manifold, effectively
suppressing the corresponding latent feature activations and
achieving targeted semantic erasure.

\textbf{2) Retouch: Semantic Completion via Positive-Guided Refinement.}
The \textit{Retouch} phase restores structural coherence while
injecting the enhanced concepts $\mathcal{D}_{\mathrm{pos}}$,
spanning the re-noising and re-denoising trajectory
$z_{t-1}' \rightarrow z_{t}^{*} \rightarrow
z_{t-1}^{*}$. We construct an enhanced prompt
$c^{+} = c \cup \mathcal{D}_{\mathrm{pos}}$ and adopt it as the
conditional input throughout both sub-steps, enabling progressive
semantic accumulation following~\cite{bai2024zigzag}.
For re-noising, we follow the same inversion procedure as
\textit{Erase} (Eqs.~\eqref{eq:erase_clean}--\eqref{eq:erase_step}),
but replace the original prompt $c$ with $c^{+}$ in the guided
noise estimate:
\begin{equation}
\label{eq:retouch_inv_guidance}
\hat{\epsilon}_{\mathrm{e}}(z_{t-1}'; t{-}1)
\;=\;
(1+\gamma_{\mathrm{e}})\,
\epsilon_\phi(z_{t-1}'; t{-}1,\, c^{+})
\;-\;
\gamma_{\mathrm{e}}\,
\epsilon_\phi(z_{t-1}'; t{-}1,\,
\mathcal{D}_{\mathrm{neg}})
\end{equation}
The re-noised state $z_{t}^{*}$ is then obtained via
Eqs.~\eqref{eq:erase_clean} and~\eqref{eq:erase_step}.
For re-denoising, starting from $z_{t}^{*}$, we apply
standard CFG (Eq.~\eqref{eq:cfg}) with $c^{+}$ and a retouch
guidance scale $\gamma_{\mathrm{r}}$:
\begin{equation}
\label{eq:retouch_guidance}
\hat{\epsilon}_{\mathrm{r}}(z_{t}^{*}; t)
\;=\;
(1+\gamma_{\mathrm{r}})\,
\epsilon_\phi(z_{t}^{*}; t,\, c^{+})
\;-\;
\gamma_{\mathrm{r}}\,
\epsilon_\phi(z_{t}^{*}; t,\, \varnothing)
\end{equation}
where $\varnothing$ denotes the unconditional input. The refined
state $z_{t-1}^{*}$ is obtained via the DDIM denoising
rule:

\begin{equation}
\label{eq:retouch_step}
z_{t-1}^{*}
=
\sqrt{\bar{\alpha}_{t-1}}
\left(
\frac{z_{t}^{*} - \sqrt{1-\bar{\alpha}_{t}}\;
\hat{\epsilon}_{\mathrm{r}}}
{\sqrt{\bar{\alpha}_{t}}}
\right)
+
\sqrt{1-\bar{\alpha}_{t-1}}\;
\hat{\epsilon}_{\mathrm{r}}
\end{equation}
By applying $c^{+}$ in both sub-steps, the positive semantic signal
is progressively accumulated into the latent representation,
completing the missing content within the erased regions.

After iterating over all inspection timesteps, the final image is
obtained by decoding ${z}_{0}^{*}$ as
${x}^{*}=\mathcal{D}_{\mathrm{vae}}({z}_{0}^{*})$.

\section{Experiments}

\subsection{Experimental Setting} \label{ex.setting}

\noindent \textbf{Implementation Details.}
We adopt two representative diffusion models in our main experiments:
the UNet-based SDXL~\cite{podell2024sdxl} and the Transformer-based
HunyuanDiT-v1.2~\cite{li2024hunyuandit}. Unless otherwise specified,
we use the DDIM sampler with $T{=}50$ denoising steps and CFG scale
$\gamma{=}5$ for both models, following their default configurations.
For \emph{Semantic Anchoring}, we leverage OpenCLIP
ViT-B/32~\cite{ilharco2021openclip} to extract semantic
priors via DAS, followed by LP-SDS
refinement with $N_{\mathrm{lp}}{=}400$. 
For \emph{Reflective Steering}, we employ Qwen-VL-Chat~\cite{bai2023qwen} as the representative VLM, with inspection stride $\tau{=}1$ and guidance scales $\gamma_{\mathrm{e}}{=}1.0$, $\gamma_{\mathrm{r}}{=}5.0$.
All experiments are conducted on NVIDIA H100 GPUs.

\noindent \textbf{Datasets and Metrics.} We conduct experiments on two benchmarks: GenEval~\cite{ghosh2023geneval} and T2I-CompBench++~\cite{huang2023t2icompbench,huang2025t2icompbench++}, to evaluate text-to-image alignment.
GenEval specializes in fine-grained object-centric metrics  and T2I-CompBench++ targets more challenging compositional scenarios.
Besides alignment, we also evaluate image quality and sample diversity, including HPSv2~\cite{wu2023human}, PickScore~\cite{kirstain2023pick}, ImageReward~\cite{xu2023imagereward}, and Aesthetic score~\cite{schuhmann2022laionb} for image quality, and LPIPS~\cite{8578166} and DINO~\cite{caron2021emerging,oquab2024dinov} similarity for diversity.
For all experiments, we follow the official benchmark protocols. Additional details of the benchmarks and evaluation metrics are provided in the supplementary material.

\subsection{Main Results} \label{main_results}

\subsubsection{Cross-Architecture Effectiveness.}

\begin{table}[t]
\centering
\caption{\textbf{Cross-architecture evaluation on GenEval.} AnchorSteer consistently improves faithfulness across  different architectures.}
\resizebox{\columnwidth}{!}{
\begin{tabular}{@{}ll ccccccc@{}}
\toprule
\textbf{Model} & \textbf{Method} & \textbf{Single}$\uparrow$ & \textbf{Two}$\uparrow$ & \textbf{Count}$\uparrow$ & \textbf{Color}$\uparrow$ & \textbf{Pos.}$\uparrow$ & \textbf{Attr.}$\uparrow$ & \textbf{Overall}$\uparrow$ \\
\midrule

\multirow{3}{*}{\textit{SDXL}}
& Standard
& 99.69 & 81.06 & 30.31 & 92.29 & 11.00 & 21.25 & 55.933 \\

& \cellcolor{blue!6}\textbf{+ AnchorSteer}
& \cellcolor{blue!6}\textbf{100.00} & \cellcolor{blue!6}\textbf{91.16} & \cellcolor{blue!6}\textbf{52.50} & \cellcolor{blue!6}\textbf{94.95} & \cellcolor{blue!6}\textbf{17.75} & \cellcolor{blue!6}\textbf{24.00} & \cellcolor{blue!6}\textbf{63.393} \\

& \multicolumn{1}{r}{\scriptsize\color{teal}$\Delta$}
& \scriptsize\color{teal}+0.31 & \scriptsize\color{teal}+10.10 & \scriptsize\color{teal}+22.19 & \scriptsize\color{teal}+2.66 & \scriptsize\color{teal}+6.75 & \scriptsize\color{teal}+2.75 & \scriptsize\color{teal}+7.460 \\

\midrule

\multirow{3}{*}{\textit{HunyuanDiT}}
& Standard
& 99.69 & 93.18 & 64.69 & 95.74 & 17.75 & 39.00 & 68.342 \\

& \cellcolor{blue!6}\textbf{+ AnchorSteer}
& \cellcolor{blue!6}\textbf{100.00} & \cellcolor{blue!6}\textbf{95.71} & \cellcolor{blue!6}\textbf{72.50} & \cellcolor{blue!6}\textbf{97.61} & \cellcolor{blue!6}\textbf{20.75} & \cellcolor{blue!6}\textbf{48.25} & \cellcolor{blue!6}\textbf{72.470} \\

& \multicolumn{1}{r}{\scriptsize\color{teal}$\Delta$}
& \scriptsize\color{teal}+0.31 & \scriptsize\color{teal}+2.53 & \scriptsize\color{teal}+7.81 & \scriptsize\color{teal}+1.87 & \scriptsize\color{teal}+3.00 & \scriptsize\color{teal}+9.25 & \scriptsize\color{teal}+4.128 \\

\bottomrule
\end{tabular}
}
\label{tab:cross_arch_geneval}
\end{table}

We evaluate AnchorSteer on two representative diffusion models with distinct architectures: 
SDXL~\cite{podell2024sdxl} and HunyuanDiT~\cite{li2024hunyuandit}. 
As shown in Table~\ref{tab:cross_arch_geneval}, AnchorSteer consistently improves the standard inference baseline across all six GenEval categories on both models.

On SDXL, AnchorSteer increases the overall score from 55.93 to 63.39, with large improvements in \textit{Counting} and \textit{Two Objects}. 
On HunyuanDiT, which already achieves a stronger baseline of 68.34, AnchorSteer raises the overall score to 72.47, with notable gains in \textit{Attribute Binding} and \textit{Counting}. 
These results demonstrate that AnchorSteer is architecture-agnostic and consistently enhances compositional faithfulness across both U-Net and Transformer backbones.

\subsubsection{Comparison with Existing Methods.}

\begin{table}[t]
\centering
\caption{\textbf{Quantitative comparison on GenEval (SDXL).} AnchorSteer achieves the best overall score and outperforms baselines on most compositional categories.}
\resizebox{\columnwidth}{!}{
\begin{tabular}{@{}l ccccccc@{}}
\toprule
\textbf{Method} & \textbf{Single}$\uparrow$ & \textbf{Two}$\uparrow$ & \textbf{Count}$\uparrow$ & \textbf{Color}$\uparrow$ & \textbf{Pos.}$\uparrow$ & \textbf{Attr.}$\uparrow$ & \textbf{Overall}$\uparrow$ \\
\midrule

Standard
& 99.69 & 81.06 & 30.31 & 92.29 & 11.00 & 21.25 & 55.933 \\

AaE~\cite{chefer2023attend}
& 99.06 & 83.59 & 31.56 & 90.96 & 15.00 & 24.00 & 57.361 \\

InitNO~\cite{guo2024initno}
& 99.69 & 85.10 & 37.81 & 89.89 & 16.00 & 22.75 & 58.540 \\

DNO~\cite{tang2025inferencetime}
& 100.00 & 80.81 & 44.06 & 90.96 & 12.75 & 22.00 & 58.430 \\

Zigzag~\cite{bai2024zigzag}
& 99.69 & 85.35 & 40.94 & 92.55 & 12.25 & 20.50 & 58.547 \\

Diffusion DPO~\cite{wallace2024diffusion}
& 100.00 & \textbf{92.42} & 46.25 & 92.29 & 15.00 & 18.50 & 60.744 \\

SPO~\cite{liang2025aesthetic}
& 100.00 & 85.86 & 33.44 & 90.96 & 13.00 & 18.50 & 56.959 \\

NPNet~\cite{zhou2025golden}
& 99.38 & 84.09 & 36.88 & 92.82 & 13.50 & 22.00 & 58.110 \\

\cellcolor{blue!6}\textbf{AnchorSteer (Ours)}
& \cellcolor{blue!6}\textbf{100.00}
& \cellcolor{blue!6}91.16
& \cellcolor{blue!6}\textbf{52.50}
& \cellcolor{blue!6}\textbf{94.95}
& \cellcolor{blue!6}\textbf{17.75}
& \cellcolor{blue!6}\textbf{24.00}
& \cellcolor{blue!6}\textbf{63.393} \\

\bottomrule
\end{tabular}
}
\label{tab:comparison_existing_methods}
\end{table}

\begin{table}[t]
\centering
\caption{\textbf{Quantitative comparison on T2I-CompBench++ (SDXL).} AnchorSteer improves compositional alignment across diverse semantic categories.}
\resizebox{\columnwidth}{!}{
\begin{tabular}{@{}l cccccccc@{}}
\toprule
\textbf{Method} & \textbf{Color}$\uparrow$ & \textbf{Shape}$\uparrow$ & \textbf{Texture}$\uparrow$ & \textbf{2D-S}$\uparrow$ & \textbf{3D-S}$\uparrow$ & \textbf{Non-S}$\uparrow$ & \textbf{Num}$\uparrow$ & \textbf{Complex}$\uparrow$ \\
\midrule

Standard
& 0.5766 & 0.4830 & 0.5353 & 0.2015 & 0.3824 & 0.3139 & 0.5029 & 0.3382 \\

AaE~\cite{chefer2023attend}
& 0.6084 & 0.4991 & 0.5511 & 0.1743 & 0.3673 & 0.3126 & 0.5017 & 0.3400 \\

InitNO~\cite{guo2024initno}
& 0.5636 & 0.4967 & 0.5430 & 0.1968 & 0.3699 & 0.3113 & 0.5037 & 0.3388 \\

DNO~\cite{tang2025inferencetime}
& 0.5982 & 0.4986 & 0.5402 & 0.2050 & 0.4024 & 0.3144 & 0.5185 & 0.3446 \\

Zigzag~\cite{bai2024zigzag}
& 0.6172 & 0.5187 & 0.5759 & 0.2011 & 0.4024 & \textbf{0.3206} & 0.5349 & 0.3504 \\

Diffusion DPO~\cite{wallace2024diffusion}
& 0.6351 & 0.5291 & 0.6110 & 0.2153 & 0.4017 & 0.3156 & 0.5223 & 0.3536 \\

SPO~\cite{liang2025aesthetic}
& 0.6194 & 0.4955 & 0.5553 & 0.1930 & 0.3928 & 0.3084 & 0.5061 & 0.3556 \\

NPNet~\cite{zhou2025golden}
& 0.5849 & 0.4824 & 0.5348 & 0.1905 & 0.3876 & 0.3132 & 0.5126 & 0.3397 \\

\cellcolor{blue!6}\textbf{AnchorSteer (Ours)}
& \cellcolor{blue!6}\textbf{0.6712}
& \cellcolor{blue!6}\textbf{0.5345}
& \cellcolor{blue!6}\textbf{0.6369}
& \cellcolor{blue!6}\textbf{0.2158}
& \cellcolor{blue!6}\textbf{0.4135}
& \cellcolor{blue!6}0.3171
& \cellcolor{blue!6}\textbf{0.5796}
& \cellcolor{blue!6}\textbf{0.3759} \\

\bottomrule
\end{tabular}
}
\label{tab:compbench}
\end{table}

Table~\ref{tab:comparison_existing_methods} presents quantitative comparisons with several representative  training-based and training-free  alignment methods on GenEval benchmark. 
AnchorSteer achieves the best overall performance with a score of 63.39, outperforming all baselines by a clear margin. 
It shows notable gains on challenging categories such as \textit{Counting}, \textit{Spatial Relation}, and \textit{Attribute Binding}, indicating stronger compositional accuracy. 

Table~\ref{tab:compbench} further evaluates AnchorSteer on T2I-CompBench++, where it achieves the best performance on the majority of semantic categories. 
These results demonstrate that AnchorSteer consistently improves compositional faithfulness and text-image alignment compared with existing methods.

\subsubsection{Qualitative Analysis.}

\begin{figure}[t]
\centering
\includegraphics[width=\linewidth]{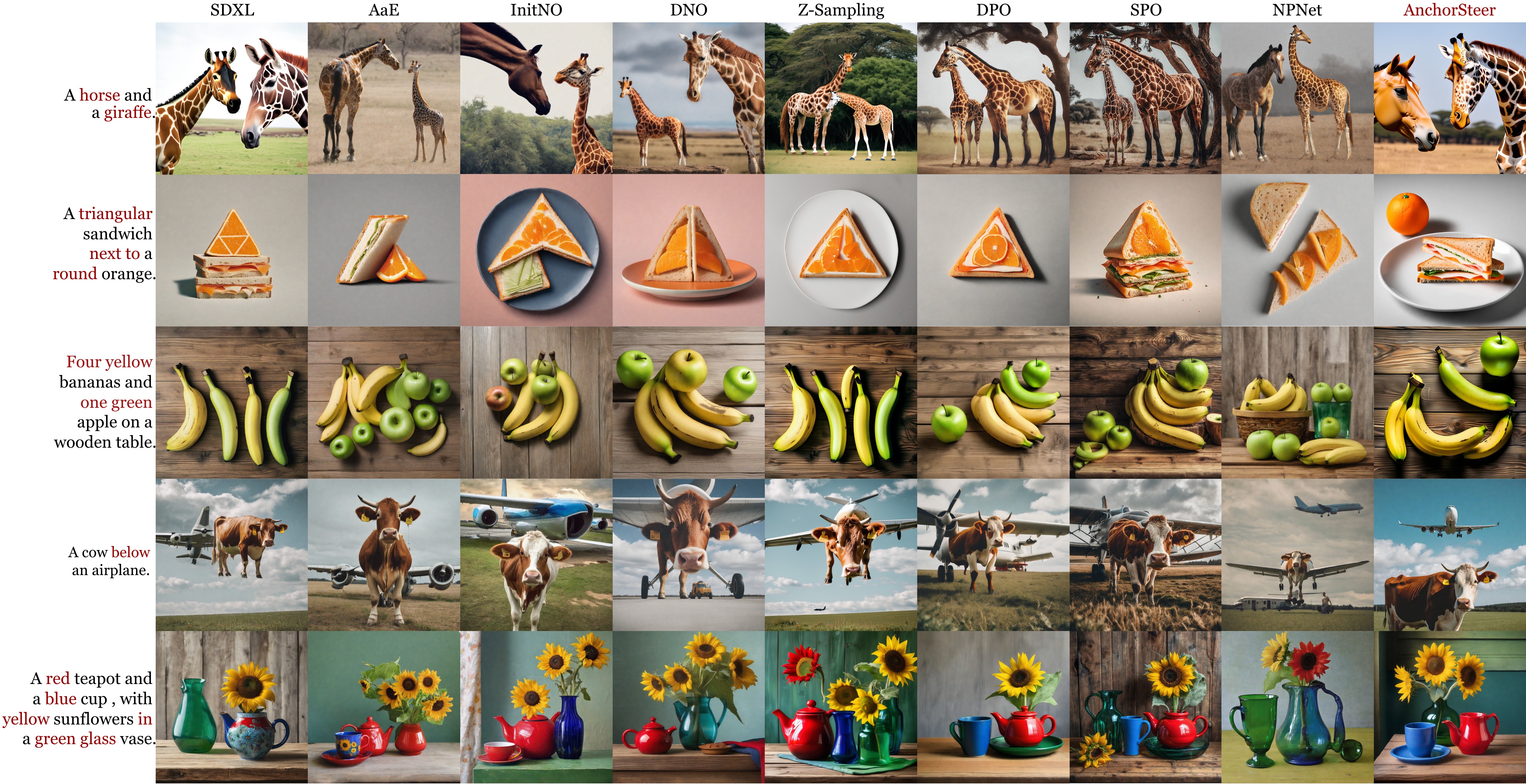}
\caption{
\textbf{Qualitative comparison on challenging compositional prompts.} AnchorSteer produces more prompt-faithful results than SDXL and several representative alignment methods, which often exhibit semantic inconsistencies.
}
\label{fig:qualitative_com}
\end{figure}

Figure~\ref{fig:qualitative_com} presents qualitative comparisons on a set of challenging compositional prompts constructed for evaluation. 
The baseline SDXL model and several representative alignment methods often exhibit semantic inconsistencies, such as missing objects, incorrect counts, attribute leakage, or violated spatial relations. 
In contrast, \textbf{AnchorSteer} produces results that more faithfully follow the prompt semantics, demonstrating improved compositional reasoning and text-image alignment.

\begin{figure}[t]
\centering
\includegraphics[width=\linewidth]{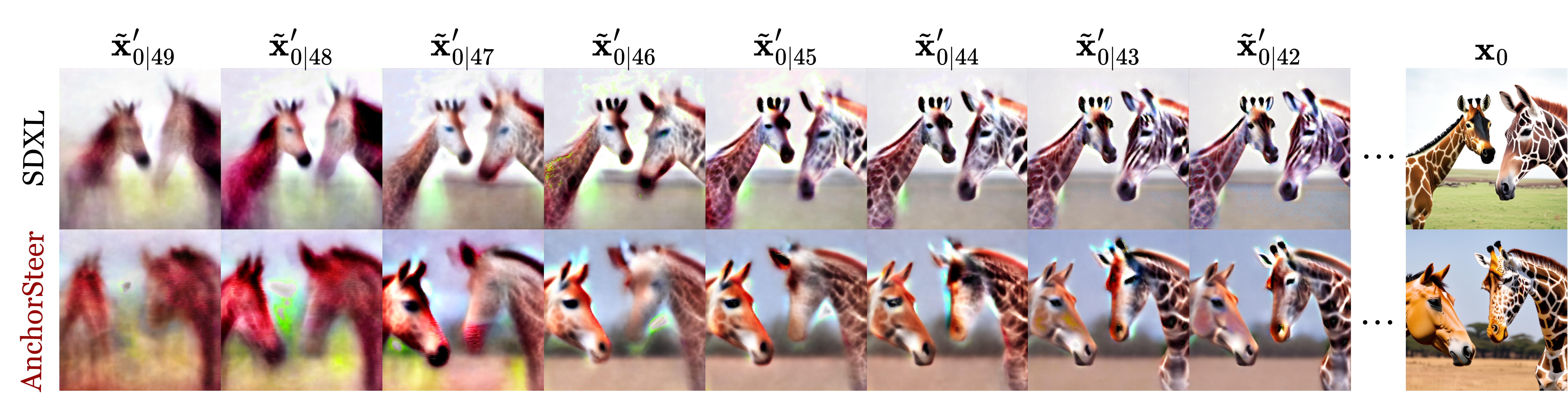}
\caption{
\textbf{Visualization of semantic correction during denoising.}
Predicted $x_0$ from the first eight steps for \textit{A horse and a giraffe}. SDXL confuses the two concepts and generates two giraffes, while AnchorSteer progressively removes the incorrect giraffe and restores the missing horse.
}
\label{fig:erase_process}
\end{figure}

Figure~\ref{fig:erase_process} further illustrates how AnchorSteer corrects semantic errors during generation. 
We visualize the predicted clean images $x_0$ from the first eight denoising steps. 
While SDXL consistently produces two giraffes, AnchorSteer gradually erases the incorrect giraffe instance and introduces the missing horse, leading to a semantically consistent result. Additional qualitative results are provided in the supplementary material.

\subsubsection{Image Quality and Diversity Analysis.}

\begin{table}[t]
\centering
\caption{\textbf{Quality and diversity analysis on GenEval(SDXL).} AnchorSteer further enhances visual quality and sample diversity over the standard baseline.}
\resizebox{\columnwidth}{!}{
\begin{tabular}{@{}l cccc|cc@{}}
\toprule
\textbf{Method} 
& \multicolumn{4}{c|}{\textbf{Image Quality}} 
& \multicolumn{2}{c}{\textbf{Sample Diversity}} \\
\cmidrule(lr){2-5} \cmidrule(l){6-7}
& \textbf{HPSv2}$\uparrow$ 
& \textbf{PickScore}$\uparrow$ 
& \textbf{ImageReward}$\uparrow$ 
& \textbf{Aesthetic}$\uparrow$
& \textbf{1-LPIPS}$\downarrow$ 
& \textbf{DINOSim}$\downarrow$ \\
\midrule

Standard
& 0.2769 & 22.55 & 0.4891 & 5.465
& 0.3860 & 0.6638 \\

\cellcolor{blue!6}\textbf{AnchorSteer (Ours)}
& \cellcolor{blue!6}\textbf{0.2963}
& \cellcolor{blue!6}\textbf{22.83}
& \cellcolor{blue!6}\textbf{0.8781}
& \cellcolor{blue!6}\textbf{5.561}
& \cellcolor{blue!6}\textbf{0.3794}
& \cellcolor{blue!6}\textbf{0.6314} \\

\bottomrule
\end{tabular}
}
\label{tab:quality_diversity}
\end{table}

We further evaluate the visual quality and sample diversity of the standard baseline and AnchorSteer.
As shown in Table~\ref{tab:quality_diversity}, AnchorSteer improves all quality and diversity metrics over the baseline, with a particularly notable gain on ImageReward~\cite{xu2023imagereward} from 0.4891 to 0.8781.
These results indicate that the improved text-image alignment is accompanied by better perceptual quality and sample diversity, rather than at their expense.
Supplementary Human preference results further support this observation.

\subsection{Ablation Studies} \label{ablation}

\subsubsection{Component-wise Analysis.}

\begin{table}[t]
\centering
\caption{\textbf{Component-wise ablation on GenEval.} Each row progressively adds one module to the vanilla baseline.}
\resizebox{\columnwidth}{!}{
\begin{tabular}{@{}ll ccccccc@{}}
\toprule
\textbf{Model} & \textbf{Method} & \textbf{Single}$\uparrow$ & \textbf{Two}$\uparrow$ & \textbf{Count}$\uparrow$ & \textbf{Color}$\uparrow$ & \textbf{Pos.}$\uparrow$ & \textbf{Attr.}$\uparrow$ & \textbf{Overall}$\uparrow$ \\
\midrule

\multirow{4}{*}{\textit{SDXL}}
& Standard
& 99.69 & 81.06 & 30.31 & 92.29 & 11.00 & 21.25 & 55.933 \\

& + Anchor
& \textbf{100.00} & 88.64 & 44.38 & 93.62 & 13.75 & \textbf{24.00} & 60.731 \\

& + Steer
& \textbf{100.00} & \textbf{91.67} & 42.50 & 92.29 & 12.50 & 20.25 & 59.868 \\

& + AnchorSteer
& \textbf{100.00} & 91.16 & \textbf{52.50} & \textbf{94.95} & \textbf{17.75} & \textbf{24.00} & \textbf{63.393} \\

\midrule

\multirow{4}{*}{\textit{HunyuanDiT}}
& Standard
& 99.69 & 93.18 & 64.69 & 95.74 & 17.75 & 39.00 & 68.342 \\

& + Anchor
& \textbf{100.00} & 93.43 & 66.88 & 96.81 & 18.50 & 46.75 & 70.395 \\

& + Steer
& \textbf{100.00} & \textbf{96.72} & 71.56 & 97.34 & 19.25 & 45.50 & 71.728 \\

& + AnchorSteer
& \textbf{100.00} & 95.71 & \textbf{72.50} & \textbf{97.61} & \textbf{20.75} & \textbf{48.25} & \textbf{72.470} \\

\bottomrule
\end{tabular}
}
\label{tab:ablation_component}
\end{table}

We progressively integrate Semantic Anchoring and Reflective Steering into the standard diffusion baseline to evaluate their individual and joint effects (Table~\ref{tab:ablation_component}). 
Semantic Anchoring consistently improves structural and attribute-related metrics, particularly \textit{Counting} and \textit{Attribute Binding}, suggesting that semantically aligned initialization reduces early-stage semantic drift. 
Reflective Steering mainly enhances relational reasoning during denoising, leading to clear improvements on the \textit{Two Objects} category and further strengthening overall compositional consistency. 
When combined, AnchorSteer achieves the best overall performance on both SDXL and HunyuanDiT, demonstrating that jointly controlling the initialization and the denoising trajectory provides complementary benefits.

\subsubsection{Effect of VLM Scale.}

\begin{table}[t]
\centering
\caption{\textbf{Effect of VLM scale on GenEval (SDXL).} We replace the VLM used for semantic diagnosis in Reflective Steering with models of different scales.}
\label{tab:vlm_scale}
\setlength{\tabcolsep}{5pt}
\renewcommand{\arraystretch}{1.05}
\begin{tabular}{@{}lcccc@{}}
\toprule
\textbf{VLM} & \textbf{Standard} & \textbf{Qwen2-VL-2B} & \textbf{Qwen2.5-VL-3B} & \textbf{Qwen-VL-Chat} \\
\midrule
\textbf{Scale} & -- & 2B & 3B & 9.6B \\
\textbf{Overall} & 55.933 & 62.073 & 62.843 & \textbf{63.393} \\
\bottomrule
\end{tabular}
\end{table}

We further examine the effect of VLM scale in Reflective Steering by comparing Qwen2-VL-2B-Instruct~\cite{Qwen2VL}, Qwen2.5-VL-3B-Instruct~\cite{qwen2.5-VL}, and Qwen-VL-Chat~\cite{bai2023qwen}. As shown in Table~\ref{tab:vlm_scale}, smaller VLMs still substantially improve over the standard SDXL baseline and achieve stronger performance than the compared baselines in Table~\ref{tab:comparison_existing_methods}, indicating that AnchorSteer does not rely on an oversized VLM for semantic diagnosis.

\begin{table}[!htbp]
\centering
\caption{\textbf{Runtime analysis.} Average per-image runtime and ablation of AnchorSteer. Diffusion DPO, SPO, and NPNet share the same inference runtime as Standard.}
\label{tab:runtime}
\setlength{\tabcolsep}{2pt}
\renewcommand{\arraystretch}{1.05}
\begin{tabular}{@{}lcccccc|cccc@{}}
\toprule
& \multicolumn{6}{c|}{\textbf{Runtime Comparison}} & \multicolumn{4}{c}{\textbf{Runtime Ablation}} \\
\cmidrule(lr){2-7} \cmidrule(l){8-11}
& Standard & Zigzag & AaE & InitNO & Ours & DNO & Anchor & Steer & Other & Total \\
\midrule
\textbf{Time (s)} & 3.7 & 8.3 & 14.9 & 34.7 & 68.5 & 155.5 & 18.0 & 46.7 & 3.8 & 68.5 \\
\bottomrule
\end{tabular}
\end{table}

\subsection{Efficiency Analysis}

Table~\ref{tab:runtime} reports the average per-image runtime of the methods evaluated in Table~\ref{tab:comparison_existing_methods}. AnchorSteer takes a runtime of 68.5 seconds per image, which is slower than standard sampling and lightweight inference-time methods, yet remains more efficient than DNO. The runtime ablation reveals that  the primary computational overhead stems from the Reflective Steering module: under the 50-step SDXL setting adopted for fair comparison, the steering loop is executed at every denoising step.  To improve efficiency and practicality, we further apply AnchorSteer to few-step distilled models, which provides a promising direction for improving efficiency without sacrificing alignment quality. More details are provided in the supplementary material.

\section{Conclusion, Limitations and Future Work}

\noindent\textbf{Limitations.}
The main limitation of AnchorSteer is its additional inference cost. 
Both Semantic Anchoring and Reflective Steering introduce extra computation compared with standard sampling, making AnchorSteer more suitable for scenarios where alignment faithfulness is prioritized over generation speed. 
Improving its efficiency while maintaining strong alignment remains an important direction for future work.

\noindent\textbf{Conclusion.}
In this work, we present AnchorSteer, a training-free framework for improving T2I alignment under complex compositional prompts. 
By jointly refining the initial noise and denoising trajectory, AnchorSteer consistently enhances compositional faithfulness across different diffusion backbones while preserving visual quality. 
Overall, AnchorSteer offers a practical inference-time solution for more faithful text-to-image generation.

\section*{Acknowledgements}
This work was supported in part by the United Fund of National Key Laboratory of Automatic Target Recognition (Shanghai) under Grant ATR(S)2025-006.

%
%

\bibliographystyle{splncs04}
\bibliography{main}

\clearpage
\appendix

\vspace*{1cm}
\begin{center}
    {\Large \textbf{Supplementary Material}}
\end{center}
\vspace{1cm}

\setcounter{figure}{0}
\setcounter{table}{0}
\setcounter{equation}{0}
\setcounter{algorithm}{0}

\renewcommand{\thefigure}{S\arabic{figure}}
\renewcommand{\thetable}{S\arabic{table}}
\renewcommand{\theequation}{S\arabic{equation}}
\renewcommand{\thealgorithm}{S\arabic{algorithm}}

This supplementary material provides additional details and experimental results to complement the main paper.

\section{Experimental Setup}
\label{sec:appendix_exp}

In this section, we provide additional details on the benchmarks, backbone models, and baseline methods used in our experiments.

\subsection{Benchmarks}

We evaluate on two benchmarks: GenEval~\cite{ghosh2023geneval} and T2I-CompBench++~\cite{huang2023t2icompbench,huang2025t2icompbench++}.

\vspace{-1.5em}
\subsubsection{GenEval.}
Compositional text–image alignment is evaluated using the GenEval benchmark, an object-centric benchmark consisting of 553 prompts covering six compositional tasks:

\vspace{-0.5em}
\begin{itemize}
\item \textbf{Single Object}: generating a specified object.
\item \textbf{Two Objects}: generating two different objects simultaneously.
\item \textbf{Counting}: generating a specified number of objects.
\item \textbf{Colors}: generating an object with a specified color.
\item \textbf{Position}: generating two objects with a specified spatial relation.
\item \textbf{Attribute Binding}: generating two objects with different color attributes.
\end{itemize}

\vspace{-0.5em}
All prompts are constructed from task-specific templates filled with randomly sampled object categories, colors, numbers, and spatial relations and we generate four images per prompt following the official protocol. For each task, we compute the accuracy over all generated images corresponding
to its prompts. The final \textbf{Overall} score is obtained by averaging the six task
accuracies.

\vspace{-0.5em}

\subsubsection{T2I-CompBench++.}
Compositional text–image alignment is further evaluated using the T2I-CompBench++ benchmark, a large-scale benchmark for fine-grained compositional text-to-image generation.

We conduct evaluation on its official validation split, which contains 2,394 prompts covering eight compositional sub-categories, including attribute binding (color, shape, and texture), spatial relationships (2D and 3D spatial relations), non-spatial relationships, numeracy, and complex compositions. Category-wise results are computed over all generated images in each sub-category following the official evaluation protocol.

\subsection{Backbone Models}

Two representative diffusion models are used in our experiments: the U-Net based SDXL~\cite{podell2024sdxl} and the Transformer-based HunyuanDiT-v1.2~\cite{li2024hunyuandit}.

\textbf{SDXL.}
SDXL is a large-scale latent diffusion model built upon a U-Net architecture. It employs a two-stage text encoder design and operates in a high-resolution latent space, enabling the generation of detailed and semantically consistent images. Due to its strong generation performance and widespread adoption, SDXL is commonly used as a benchmark backbone in text-to-image generation research.

\textbf{HunyuanDiT-v1.2.}
HunyuanDiT is a diffusion model based on the Diffusion Transformer architecture, where Transformer blocks process latent tokens instead of convolutional feature maps. By leveraging self-attention over latent representations, the model can capture global dependencies within the generated image. HunyuanDiT-v1.2 is a recent large-scale text-to-image diffusion model that provides competitive generation performance.

Unless otherwise specified, all experiments are conducted under the default inference configurations of the corresponding models, as described in the main paper. AnchorSteer is applied as a training-free inference-time optimization module without modifying any model parameters.

\subsection{Baseline Methods}

We compare AnchorSteer with several representative alignment methods, including both training-free and training-based approaches.

\textbf{Attend-and-Excite}~\cite{chefer2023attend} is a training-free method that improves compositional text-to-image generation by optimizing cross-attention maps during denoising to strengthen the grounding of subject tokens in the generated image.

\textbf{InitNO}~\cite{guo2024initno} is a training-free method that improves text-to-image alignment by optimizing the initial Gaussian noise to guide it toward semantically valid regions prior to denoising.

\textbf{DNO}~\cite{tang2025inferencetime} (Direct Noise Optimization) is a training-free method that optimizes the noise vectors during diffusion sampling to maximize a reward on the generated image. PickScore~\cite{kirstain2023pick} is adopted as the reward function provided in the official implementation.

\textbf{Zigzag}~\cite{bai2024zigzag} is a training-free sampling strategy that introduces a diffusion self-reflection mechanism by alternating between denoising and inversion steps during inference, enabling semantic information to be progressively injected into the latent variables.

\textbf{Diffusion-DPO}~\cite{wallace2024diffusion} fine-tunes diffusion models using Direct Preference Optimization on human preference pairs, encouraging the model to generate preferred images relative to a frozen reference model. 

\textbf{SPO}~\cite{liang2025aesthetic} (Step-by-step Preference Optimization) constructs preference pairs at each denoising step using a step-aware preference model rather than propagating trajectory-level labels.

\textbf{NPNet}~\cite{zhou2025golden} is a learning-based method that predicts text-conditioned golden noise using a model trained on a large-scale noise-pair dataset, enabling direct generation of initial noise for diffusion sampling.

For training-based methods, we use their officially released SDXL fine-tuned checkpoints. For training-free methods, we adopt the hyperparameter settings recommended in their original papers. All methods are evaluated on the same GenEval prompts using an identical evaluation pipeline.

\subsection{Evaluation Metrics}

We evaluate both image quality and sample diversity using standard metrics.
For sample diversity, we measure pairwise similarity among multiple outputs generated from the same prompt.

\textbf{HPSv2}~\cite{wu2023human} is trained on large-scale human preference data and evaluates how well generated images align with human judgments of visual quality and prompt faithfulness.

\textbf{PickScore}~\cite{kirstain2023pick} is a learning-based reward model trained on human preference comparisons, measuring relative preference between generated images and text prompts.

\textbf{ImageReward}~\cite{xu2023imagereward} is a human feedback-based reward model that jointly evaluates realism, prompt adherence, and overall visual quality.

\textbf{Aesthetic score}~\cite{schuhmann2022laionb} is a CLIP-based aesthetic predictor trained on LAION data, estimating the perceived visual appeal of images.

\textbf{LPIPS}~\cite{8578166} measures perceptual distance between image pairs based on deep feature representations. We use $1$-LPIPS as a diversity metric, where smaller values indicate greater variation among generated samples.

\textbf{DINO similarity}~\cite{caron2021emerging,oquab2024dinov} computes cosine similarity in DINOv2 feature space to measure semantic similarity among samples; lower values indicate higher diversity.
\section{Detailed Derivation of LP-SDS}
\label{sec:appendix_lp}

This section provides the derivation of Eq.~(4) in the main paper, discusses its KL interpretation in Eq.~(5), and compares LP-SDS with standard SDS.

\textbf{Derivation of Eq.~(4).}
Let $z=\mathcal{E}_{\mathrm{VAE}}(I_{\mathrm{prior}})$ denote the latent
variable optimized in LP-SDS, initialized from the semantic prior. The forward diffusion process is
\begin{align}
z_t = \alpha_t z + \sigma_t\epsilon,
\qquad
\epsilon\sim\mathcal{N}(0,I).
\end{align}

The standard diffusion noise-prediction objective is
\begin{align}
\mathcal{L}_{\mathrm{Diff}}(\phi,z)
=
\mathbb{E}_{t,\epsilon}\!\left[
w(t)\,
\|\hat{\epsilon}_\phi(z_t;c,t)-\epsilon\|_2^2
\right].
\end{align}

Taking gradients with respect to $z$ and applying the chain rule gives
\begin{align}
\nabla_z\mathcal{L}_{\mathrm{Diff}}
=
\mathbb{E}_{t,\epsilon}\!\left[
2w(t)\,
\frac{\partial z_t}{\partial z}^{\!\top}
\frac{\partial\hat{\epsilon}_\phi(z_t;c,t)}{\partial z_t}^{\!\top}
(\hat{\epsilon}_\phi(z_t;c,t)-\epsilon)
\right].
\end{align}

Since $z_t=\alpha_t z+\sigma_t\epsilon$, we have
\begin{align}
\frac{\partial z_t}{\partial z}=\alpha_t I.
\end{align}

Substituting yields
\begin{align}
\nabla_z\mathcal{L}_{\mathrm{Diff}}
=
\mathbb{E}_{t,\epsilon}\!\left[
2\alpha_t w(t)
\left(\frac{\partial\hat{\epsilon}_\phi(z_t;c,t)}{\partial z_t}\right)^{\!\top}
(\hat{\epsilon}_\phi(z_t;c,t)-\epsilon)
\right].
\end{align}

Following Score Distillation Sampling (SDS)~\cite{poole2023dreamfusion},
the Jacobian term $\partial\hat{\epsilon}_\phi/\partial z_t$ is omitted
and all remaining scalar factors are absorbed into $w(t)$, yielding the approximate gradient
\begin{align}
\nabla_z\mathcal{L}_{\mathrm{LP\text{-}SDS}}
\triangleq
\mathbb{E}_{t,\epsilon}\!\left[
{w}(t)
(\hat{\epsilon}_\phi(z_t;c,t)-\epsilon)
\right],
\end{align}
which corresponds to Eq.~(4) in the main paper.

\textbf{KL interpretation (Eq.~(5)).}
Following the analysis in~\cite{poole2023dreamfusion}, the approximate gradient
above admits the following KL interpretation:
\begin{align}
\nabla_z\mathcal{L}_{\mathrm{LP\text{-}SDS}}
=
\nabla_z
\mathbb{E}_t\!\left[
\frac{\sigma_t}{\alpha_t}w(t)
\mathrm{KL}\!\left(
q(z_t\mid z)\,\|\,p_\phi(z_t;c,t)
\right)
\right],
\end{align}
where $q(z_t\mid z)=\mathcal{N}(\alpha_t z,\sigma_t^2 I)$ denotes the
forward diffusion distribution and $p_\phi(z_t;c,t)$ denotes the model
distribution implicitly defined by the diffusion model.

\textbf{Comparison with Standard SDS.}
Compared to standard SDS, LP-SDS introduces two key modifications. \textbf{First}, LP-SDS performs optimization directly in the VAE latent space
instead of updating the parameters of a differentiable renderer
(e.g., a NeRF) as in standard SDS.
As a result, backpropagation through an intermediate renderer or
the VAE encoder is avoided, and the optimized latent can be
directly used for subsequent diffusion sampling. \textbf{Second}, the semantic prior $I_{\mathrm{prior}}$ already provides
a meaningful initialization close to the target.
Therefore, only a small number of optimization steps
 $N_{\mathrm{lp}}$ are required to achieve sufficient alignment.
This mitigates the over-saturation artifacts often observed in SDS,
which typically arise from prolonged aggressive optimization.
\section{Complete Algorithm of AnchorSteer}
\label{sec:appendix_algo}

This section provides the pseudo-code of AnchorSteer, consisting of semantic anchoring for initialization and reflective steering during the denoising trajectory.

\begin{breakablealgorithm}
\caption{AnchorSteer Inference Algorithm}
\label{alg:main_algorithm}
\begin{algorithmic}[1]

\Input Diffusion model $\epsilon_\phi$, VAE $\mathcal{E}/\mathcal{D}$, CLIP $\mathcal{E}_I/\mathcal{E}_T$, VLM $\mathcal{V}$, prompt $c$, steps $T$, inspection set $\mathcal{T}_{\mathrm{insp}}$, scales $\gamma,\gamma_{\mathrm{e}},\gamma_{\mathrm{r}}$.

\Output Generated image $x^*$.

\Statex \textbf{// Component I: Semantic Anchoring}

\State $I_{\mathrm{prior}} \gets \mathrm{DAS}(c,\mathcal{E}_I,\mathcal{E}_T)$

\State $z^* \gets \mathrm{LP\text{-}SDS}(\mathcal{E}(I_{\mathrm{prior}}),\epsilon_\phi,c)$

\State $z_T \gets \sqrt{\bar{\alpha}_T}\,z^*
      + \sqrt{1-\bar{\alpha}_T}\,\epsilon$,
      \quad $\epsilon\sim\mathcal{N}(\mathbf{0},\mathbf{I})$

\Statex \textbf{// Component II: Reflective Steering}

\For{$t=T,\dots,1$}

    \State $\hat{\epsilon} \gets
    (1+\gamma)\epsilon_\phi(z_t;t,c)
    -\gamma\epsilon_\phi(z_t;t,\varnothing)$

    \State $z'_{t-1} \gets
    \mathrm{DDIM\text{-}Denoise}(z_t,\hat{\epsilon},t)$

    \If{$t\in\mathcal{T}_{\mathrm{insp}}$}

        \Statex \hspace{\algorithmicindent}\hspace{\algorithmicindent}
        \textit{// Think phase}

        \State $\tilde{z}'_{0|t-1} \gets
        (z'_{t-1}-\sqrt{1-\bar{\alpha}_{t-1}}\hat{\epsilon})
        /\sqrt{\bar{\alpha}_{t-1}}$

        \State $\tilde{x}'_{0|t-1} \gets
        \mathcal{D}(\tilde{z}'_{0|t-1})$

        \State $(\mathcal{D}_{\mathrm{pos}},\mathcal{D}_{\mathrm{neg}})
        \gets \mathcal{V}(\tilde{x}'_{0|t-1},c)$

        \If{$\mathcal{D}_{\mathrm{pos}}\neq\emptyset$
        \textbf{or}
        $\mathcal{D}_{\mathrm{neg}}\neq\emptyset$}

            \Statex \hspace{\algorithmicindent}\hspace{\algorithmicindent}
            \hspace{\algorithmicindent}
            \textit{// Erase and Retouch phase}

            \State $c^+ \gets c\cup\mathcal{D}_{\mathrm{pos}}$

            \State $\hat{\epsilon}_{\mathrm{e}} \gets
            (1+\gamma_{\mathrm{e}})
            \epsilon_\phi(z'_{t-1};t{-}1,c^+)
            -
            \gamma_{\mathrm{e}}
            \epsilon_\phi(z'_{t-1};t{-}1,\mathcal{D}_{\mathrm{neg}})$

            \State $\hat{z}_{0|t-1} \gets
            (z'_{t-1}-\sqrt{1-\bar{\alpha}_{t-1}}\hat{\epsilon}_{\mathrm{e}})
            /\sqrt{\bar{\alpha}_{t-1}}$

            \State $z_t^{*} \gets
            \sqrt{\bar{\alpha}_t}\hat{z}_{0|t-1}
            +\sqrt{1-\bar{\alpha}_t}\hat{\epsilon}_{\mathrm{e}}$

            \State $\hat{\epsilon}_{\mathrm{r}} \gets
            (1+\gamma_{\mathrm{r}})
            \epsilon_\phi(z_t^{*};t,c^+)
            -
            \gamma_{\mathrm{r}}
            \epsilon_\phi(z_t^{*};t,\varnothing)$

            \State $z'_{t-1} \gets
            \mathrm{DDIM\text{-}Denoise}(z_t^{*},
            \hat{\epsilon}_{\mathrm{r}},t)$

        \EndIf

    \EndIf

    \State $z_{t-1} \gets z'_{t-1}$

\EndFor

\State $x^* \gets \mathcal{D}(z_0)$
\State \Return $x^*$

\end{algorithmic}
\end{breakablealgorithm}
\section{Additional Analysis of Noise Initialization}
\label{sec:supp_noise_init}

In this section, we provide additional analysis on noise initialization, including the semantic non-equivalence of initial noise samples and the limitations of existing initialization approaches. These discussions further clarify the motivation behind our proposed Semantic Anchoring method.

\subsection{Semantic Non-equivalence of Initial Noise}

In standard diffusion sampling, the denoising trajectory starts from an initial noise $z_T \sim \mathcal{N}(0,\mathbf{I})$. For example, to generate five different images of tulips, one typically draws five independent Gaussian noise samples and runs the denoising process from each using the same text prompt, as illustrated in Figure~2a of the main paper. Intuitively, since all samples are drawn from the same isotropic Gaussian, they might be expected to behave equivalently with respect to the generation outcome.

However, recent studies~\cite{qi2024noisescreatedequallydiffusionnoise,bai2024zigzag,guo2024initno,yan2025beyond,wu2023membership} reveal that this is not the case. Empirically, some initial noises produce images that align well with the prompt even under weak classifier-free guidance (low CFG scale), whereas others still drift away from the intended semantics despite strong guidance. This indicates that certain initial noises are intrinsically more compatible with the target semantics, resulting in more prompt-faithful generations.

\subsection{Limitations of Search-Based Initialization}

One line of research attempts to improve initialization through candidate
search~\cite{10656825,
samuel2024generating,kim2026model,
ma2025inference}, as illustrated in Figure~2b of the main paper.
These methods generate a pool of noise samples and select the most promising
candidate according to a predefined scoring function.
Despite their conceptual simplicity, search-based methods suffer from three
fundamental limitations:

\textbf{(1) High computational cost.}
These methods typically require evaluating a large number of candidates,
each involving at least a partial denoising pass, making the approach
computationally expensive.

\textbf{(2) Pool-bounded search space.}
Their effectiveness is fundamentally constrained by the quality and diversity
of the noise pool, which provides no guarantee of containing a truly
prompt-aligned initialization.

\textbf{(3) Unreliable scoring functions.}
The search is usually driven by scoring functions that may not faithfully
capture nuanced text--image alignment, especially for compositional prompts.

\subsection{Limitations of Manipulation-Based Initialization}

Another category of approaches attempts to directly refine a sampled noise
into a more prompt-compatible starting point~\cite{ahn2026a,
zhou2025golden,zeng2025d2dpm,tang2025inferencetime,harrington2026s},
as illustrated in Figure~2c of the main paper. Instead of evaluating many independently
sampled noises, these methods modify a single random noise to be more
prompt-aligned. Specifically, they can be divided into two subcategories:
\emph{learning-based} methods, which train a transformation
network~\cite{ahn2026a,zhou2025golden} to map random noise into a
more prompt-compatible form, and \emph{optimization-based} methods, which
update the noise via gradient descent with a predefined reward
objective~\cite{zeng2025d2dpm,tang2025inferencetime,harrington2026s}.
Despite their effectiveness, manipulation-based methods share two primary
limitations:

\textbf{(1) Unreliable semantic injection.}
From the perspective of semantic injection, these methods cannot guarantee
that manipulation reliably embeds text-aligned semantics into the refined
noise. For learning-based approaches, the transformation network typically
relies on algebraic decompositions (e.g., SVD), yet there is no theoretical
evidence that such operations are sufficient to achieve meaningful semantic
injection~\cite{zhou2025golden}. For optimization-based approaches, it
remains difficult to construct a reward objective that accurately reflects
fine-grained text--image alignment, especially for compositional prompts
involving multiple objects, attributes, and spatial relations.

\textbf{(2) Distribution shift risk.}
From the perspective of noise properties, the manipulation may introduce
distributional shifts that undermine the denoising process. Diffusion models
are trained by corrupting real images into Gaussian noise and then learning
to reverse this corruption; the initial latent is therefore assumed to follow
$\mathcal{N}(0, \mathbf{I})$. However, existing manipulation-based methods cannot ensure that the refined noise remains Gaussian or resembles the corruption-based noisy data encountered during training, leading to out-of-distribution inputs that degrade generation quality.

\section{Supplementary Experimental Results}

In this section, we provide additional experimental results that
further support the analyses and conclusions presented in the main paper.

\subsection{Visualization of the LP-SDS Optimization Process}

\begin{figure}[t]
\centering
\includegraphics[width=\linewidth]{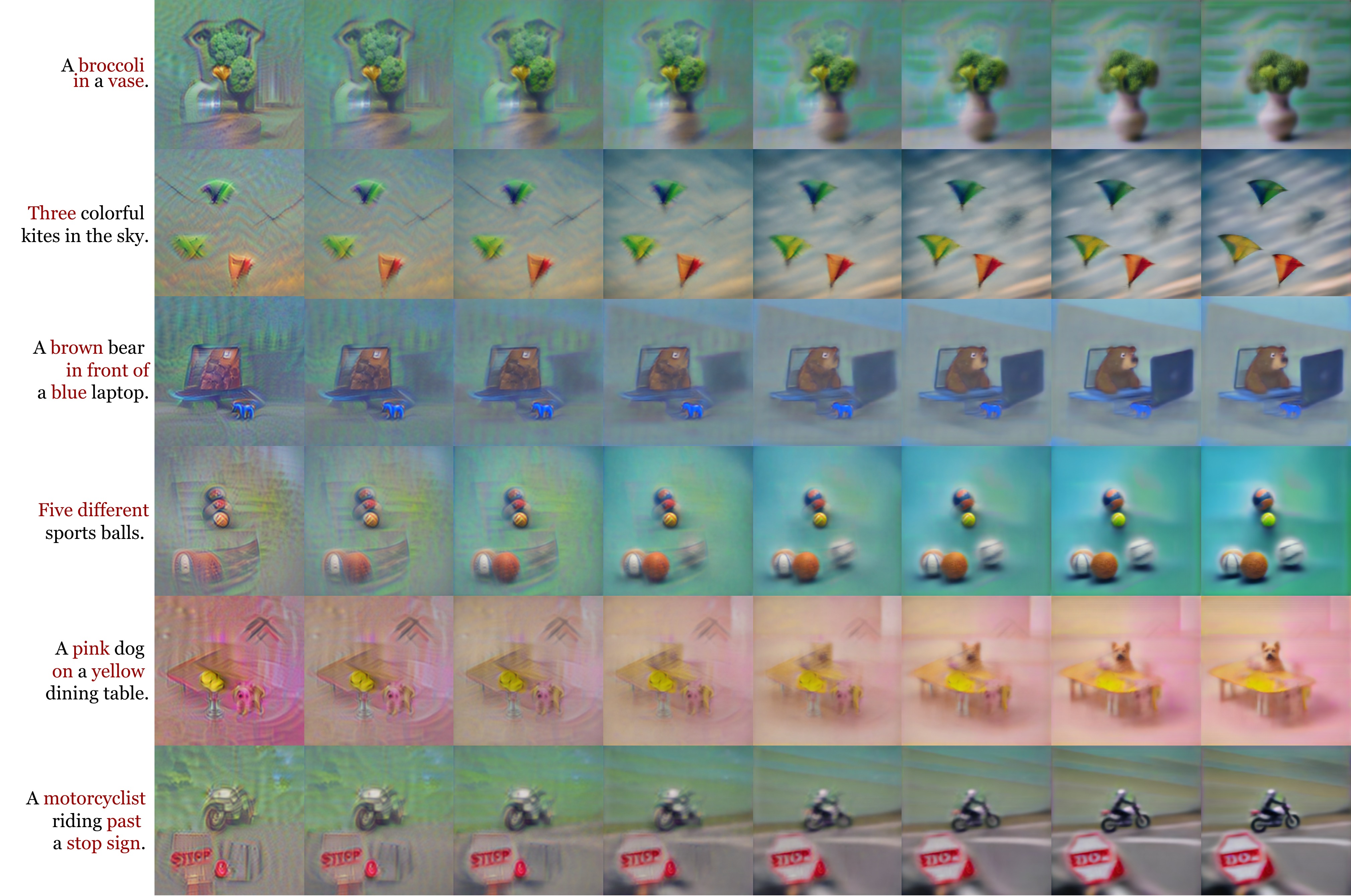}
\caption{
\textbf{Visualization of the LP-SDS optimization process.}
Each row corresponds to a different prompt, and the columns show
intermediate states during the optimization from the CLIP-based
semantic prior $I_{\mathrm{prior}}$ to the refined image $I^{*}$.
}
\label{fig:lp_optimization}
\end{figure}

Figure~\ref{fig:lp_optimization} further visualizes the optimization
process of LP-SDS on additional examples. Starting from the CLIP-based
semantic prior $I_{\mathrm{prior}}$, the latent representation is
iteratively refined using diffusion gradients. As optimization
progresses, the artifacts present in the initial CLIP synthesis
gradually disappear and the images become increasingly natural and
coherent, indicating that the latent is progressively projected onto
the diffusion model manifold while preserving the semantics specified
by the text prompt.

\subsection{Other Ablation Study}

\subsubsection{Diversity Analysis.}
We further analyze the effect of each AnchorSteer component on sample diversity. 
Specifically, we sample 20 prompts per GenEval task and generate 10 images per prompt.
As shown in Table~\ref{tab:diversity_ablation}, the Anchor component slightly reduces 
sample diversity, whereas the Steer component consistently enhances it. Overall, AnchorSteer 
achieves superior diversity compared to Standard SDXL across both metrics.

\begin{table}[t]
\centering
\caption{Ablation study of AnchorSteer components on sample diversity.}
\label{tab:diversity_ablation}
\setlength{\tabcolsep}{6pt}
\begin{tabular}{lcc}
\toprule
\textbf{Method} & \textbf{1-LPIPS} $\downarrow$ & \textbf{DINOSim} $\downarrow$ \\
\midrule
Standard       & 0.3860 & 0.6638 \\
+Anchor   & 0.4406 & 0.6755 \\
+Steer    & \textbf{0.3236} & 0.6585 \\
Ours      & 0.3794 & \textbf{0.6314} \\
\bottomrule
\end{tabular}
\end{table}

\subsubsection{Hyperparameter-wise Analysis.}

We conduct ablation experiments on two key hyperparameters:
the number of LP-SDS optimization steps $N_{\mathrm{lp}}$
and the inspection stride $\tau$.

\textbf{1) Effect of $N_{\mathrm{lp}}$.}
We vary $N_{\mathrm{lp}} \in \{0, 200, 400, 600\}$.
As shown in Table~\ref{tab:ablation_hyper} (top), the CLIP-only prior ($N_{\mathrm{lp}}{=}0$) improves the standard baseline, while applying LP-SDS optimization ($N_{\mathrm{lp}}>0$) further boosts performance.
This confirms both the effectiveness of the semantic prior and the need to bridge the domain gap between CLIP embeddings and the diffusion latent space. Performance peaks at $N_{\mathrm{lp}}{=}400$ and slightly declines at 600, likely due to mild over-saturation from prolonged optimization.

\textbf{2) Effect of Inspection Stride $\tau$.}
We test $\tau \in \{1, 2, 5\}$.
As shown in Table~\ref{tab:ablation_hyper} (bottom), smaller stride
(more frequent inspection) consistently improves alignment,
particularly on compositional categories such as \emph{Two-object},
\emph{Counting}, and \emph{Position}, where semantic deviations
are most prone to accumulate. Notably, densely applying the correction loop does not compromise generation stability, confirming
that the local bidirectional editing mechanism introduces no
significant distributional shift.

\begin{table}[t]
\centering
\caption{Ablation study of AnchorSteer hyperparameters (GenEval, SDXL).}
\footnotesize
\setlength{\tabcolsep}{4.5pt}
\begin{tabular}{@{}cl ccccccc@{}}
\toprule
& & \textbf{Single$\uparrow$}
  & \textbf{Two$\uparrow$}
  & \textbf{Count$\uparrow$}
  & \textbf{Color$\uparrow$}
  & \textbf{Pos.$\uparrow$}
  & \textbf{Attr.$\uparrow$}
  & \textbf{Overall$\uparrow$} \\
\midrule
\multirow{4}{*}{\rotatebox{90}{$N_{\mathrm{lp}}$}}
& 0   & 100.00 & 87.88 & 49.06 & 94.41 & 14.25 & 19.00 & 60.768 \\
& 200 & 100.00 & 88.38 & 49.38 & 93.88 & 14.00 & 22.75 & 61.399 \\
& 400 & \textbf{100.00} & \textbf{91.16} & \textbf{52.50} & \textbf{94.95} & \textbf{17.75} & \textbf{24.00} & \textbf{63.393} \\
& 600 & 100.00 & 90.66 & 52.50 & 94.41 & 15.75 & 21.75 & 62.512 \\
\midrule
\multirow{3}{*}{\rotatebox{90}{$\tau$}}
& 1 & \textbf{100.00} & \textbf{91.16} & \textbf{52.50} & \textbf{94.95} & \textbf{17.75} & \textbf{24.00} & \textbf{63.393} \\
& 2 & 100.00 & 87.63 & 49.06 & 94.15 & 14.75 & 21.75 & 61.223 \\
& 5 & 100.00 & 86.87 & 45.31 & 92.02 & 13.25 & 22.25 & 59.950 \\
\bottomrule
\end{tabular}
\label{tab:ablation_hyper}
\end{table}

\subsection{Additional Qualitative Results}

We provide additional qualitative comparisons between standard sampling and our AnchorSteer method across various compositional generation tasks.

\begin{figure}[t]
    \centering
    \includegraphics[width=\linewidth]{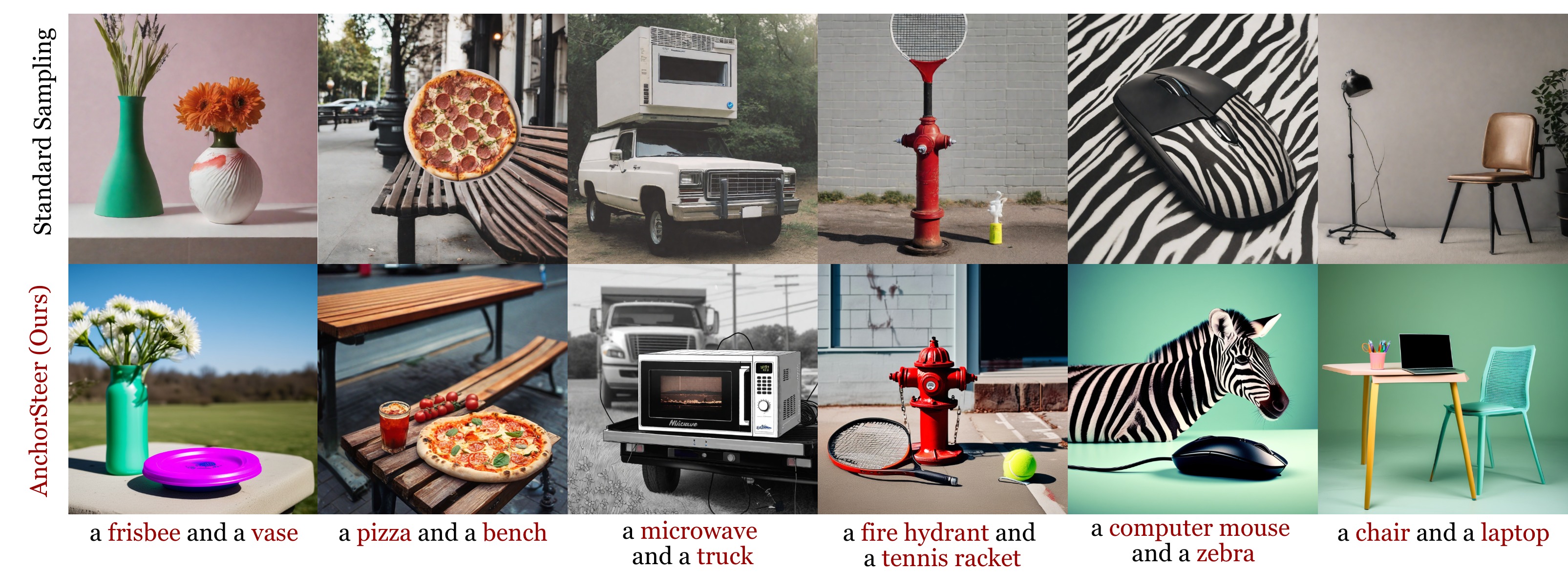}
    \caption{Qualitative comparison on \textbf{two-object generation}.}
    \label{fig:appendix_two_ob}
\end{figure}

\textbf{Two Objects.}
As shown in Figure~\ref{fig:appendix_two_ob}, standard sampling often fails to generate both objects, either missing one entirely or merging them into a single entity. In contrast, AnchorSteer consistently produces both distinct objects as specified in the prompt.

\begin{figure}[t]
    \centering
    \includegraphics[width=\linewidth]{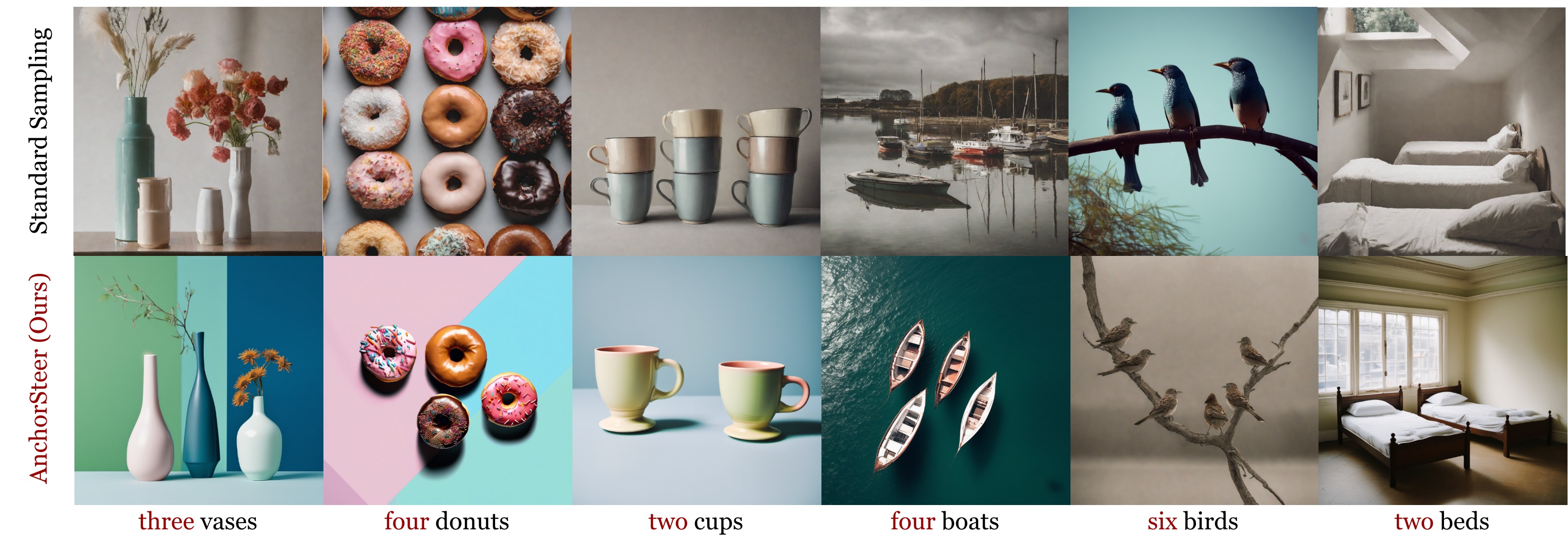}
    \caption{Qualitative comparison on \textbf{object counting}.}
    \label{fig:appendix_count}
\end{figure}

\textbf{Counting.}
Figure~\ref{fig:appendix_count} presents results on object counting prompts. Standard sampling frequently generates an incorrect number of objects, while AnchorSteer accurately produces the exact quantity specified in the text prompt.

\begin{figure}[t]
    \centering
    \includegraphics[width=\linewidth]{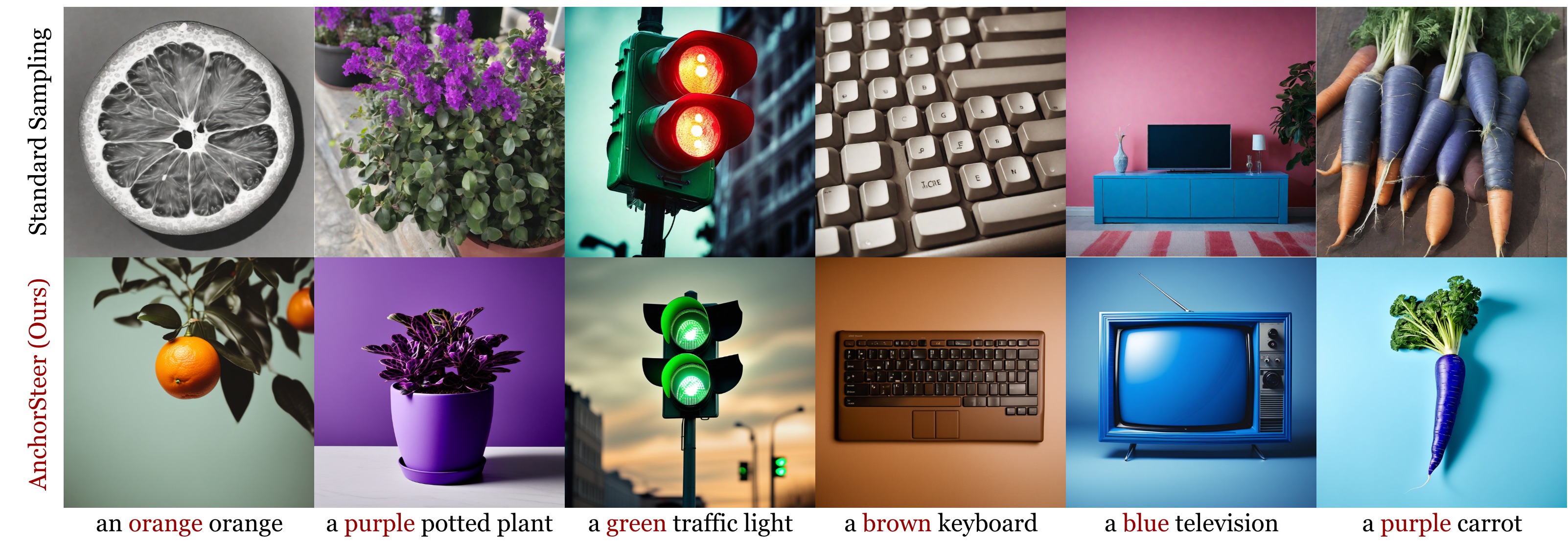}
    \caption{Qualitative comparison on \textbf{color}.}
    \label{fig:appendix_color}
\end{figure}

\textbf{Color.}
In Figure~\ref{fig:appendix_color}, we show results for prompts requiring objects with specific colors. Standard sampling sometimes generates objects with incorrect or default colors, while AnchorSteer reliably produces each object with the correct color, even for uncommon color-object combinations (\emph{e.g.}, a purple carrot).

\begin{figure}[t]
    \centering
    \includegraphics[width=\linewidth]{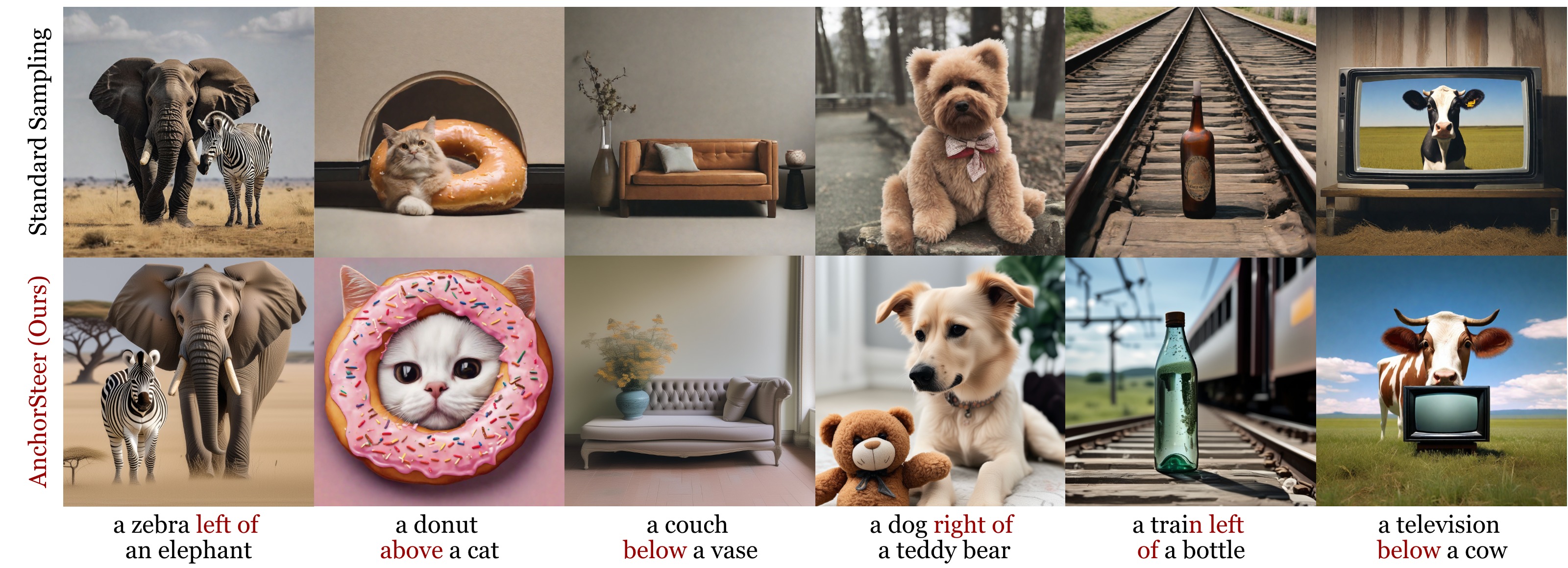}
    \caption{Qualitative comparison on \textbf{spatial relationships}.}
    \label{fig:appendix_position}
\end{figure}

\textbf{Spatial Relationships.}
Figure~\ref{fig:appendix_position} illustrates results on spatial relationship prompts. Standard sampling often places objects in incorrect relative positions, while AnchorSteer arranges them according to the specified spatial constraints.

\begin{figure}[t]
    \centering
    \includegraphics[width=\linewidth]{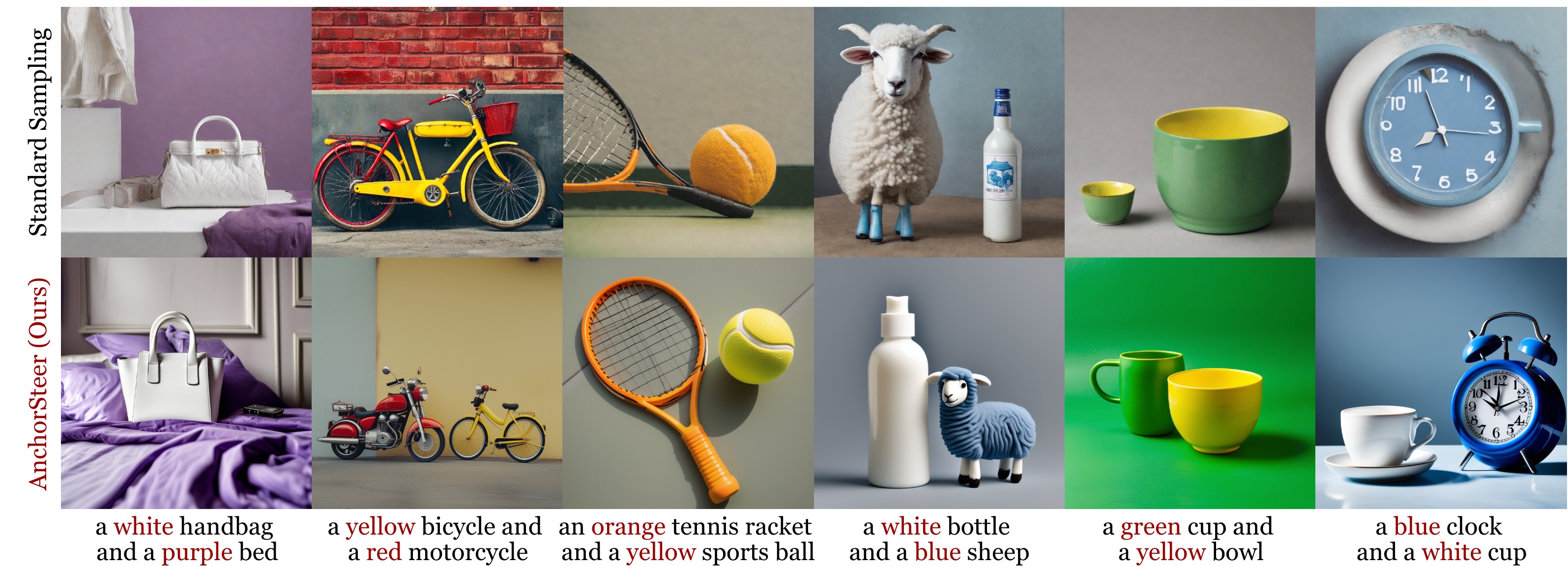}
    \caption{Qualitative comparison on \textbf{multi-object attribute binding}.}
    \label{fig:appendix_attribute}
\end{figure}

\textbf{Multi-Object Attribute\textbf Binding.}
As shown in Figure~\ref{fig:appendix_attribute}, when multiple objects with distinct attributes appear in the same prompt, standard sampling tends to swap or ignore the specified attributes. AnchorSteer correctly binds each attribute to its corresponding object, avoiding attribute leakage between objects.

\subsection{User Study}

A user study  is conducted to evaluate human preference between AnchorSteer and the SDXL baseline in terms of text-image alignment and image quality. We randomly sample prompts covering different compositional aspects, and participants are asked to select the result that better matches the text prompt and has higher visual quality, with a tie option allowed.

Table~\ref{tab:user_study} shows that AnchorSteer is preferred more often in terms of alignment with a win rate of 65.3\%. For image quality, it achieves comparable performance with a win rate of 43.5\% and a relatively high tie ratio, indicating no degradation in visual quality.

\subsection{Results on Distilled Models}

We additionally evaluate AnchorSteer on the distilled four-step diffusion model 
Qwen-Image-Lightning-4steps(Qwen-L4)~\cite{wu2025qwenimagetechnicalreport}. As shown in Table~\ref{tab:qwen_l4}, AnchorSteer achieves a higher GenEval overall score on Qwen-L4 than the full-step Qwen-Image~\cite{wu2025qwenimagetechnicalreport} baseline, while using lower inference cost, indicating the practical value of our method for efficient diffusion-based generation.

\begin{table}[t]
\centering

\begin{minipage}[t]{0.49\linewidth}
\vspace{0pt}
\centering
\caption{User study results.}
\label{tab:user_study}
\setlength{\tabcolsep}{7pt}
\begin{tabular}{lccc}
\toprule
\textbf{Criterion} & \textbf{Win} & \textbf{Tie} & \textbf{Lose} \\
\midrule
Alignment & \textbf{65.3}\% & 11.8\% & 22.9\% \\
Quality & \textbf{43.5}\% & 31.2\% & 25.3\% \\
\bottomrule
\end{tabular}
\end{minipage}
\hfill
\begin{minipage}[t]{0.49\linewidth}
\vspace{0pt}
\centering
\caption{Results on Qwen-L4.}
\label{tab:qwen_l4}
\setlength{\tabcolsep}{5.5pt}
\renewcommand{\arraystretch}{1.1}
\begin{tabular}{lcc}
\toprule
\textbf{Method} & \textbf{Overall}  & \textbf{Time (s)} \\
\midrule
Qwen-Image      & 89.902 & 26.8 \\
Qwen-L4         & 89.383 & 1.9 \\
\textbf{Ours}   & \textbf{91.293} & 24.8 \\
\bottomrule
\end{tabular}
\end{minipage}

\end{table}

\section{Semantic Preservation in $z_T$}
\label{sec:semantic_preserve}

We study whether semantic information is preserved in $z_T$ after the standard DDPM forward diffusion process.

\textbf{Theoretical analysis.}
Prior work~\cite{Everaert_2024_WACV} shows that the DDPM forward process does not fully erase image semantics, and that semantic leakage remains and has been exploited in diverse generation tasks~\cite{Everaert_2024_WACV,koo2024flexiedit,xu2025stylessp,wu2024freeinit}. Based on this insight, we use the leakage from $z^*$ as a semantic bias for improving text-image alignment.

\textbf{Experimental validation.}
Table~\ref{tab:semantic_preserve} compares different initialization strategies, including pure Gaussian noise $z_G$, shuffled latent $z_T^{shu}$ (from a mismatched prompt), and our initialized latent $z_T^{ours}$. The results show that $z_T^{ours}$ preserves more useful semantic information from $z^*$, leading to better text-image alignment.

\begin{table}[t]
\centering
\caption{\textbf{Semantic preservation in $z_T$ under DDPM forward diffusion on GenEval (SDXL).} We compare different initialization strategies: $z_G$, $z_T^{shu}$, and $z_T^{ours}$.}
\setlength{\tabcolsep}{10pt}
\begin{tabular}{@{}l ccc@{}}
\toprule
\textbf{Method} & \textbf{$z_G$} & \textbf{$z_T^{shu}$} & \textbf{$z_T^{ours}$} \\
\midrule
Overall & 59.868 & 59.850 & \textbf{63.363} \\
\bottomrule
\end{tabular}

\label{tab:semantic_preserve}
\end{table}

\section{VLM Inspection Details}

We use a predefined VLM instruction template to inspect the decoded intermediate
image against the text prompt. The full templates and examples are provided below.

\paragraph{Constraint extraction prompt.} \noindent

\noindent Input: text prompt.

\begin{verbatim}
You are a prompt decomposer for text-to-image evaluation.
Given a PROMPT, extract visually verifiable constraints as a JSON list.
RULES:
- Extract ONLY concrete visual requirements
  (objects, attributes, counts, relations).
- Do NOT extract conjunctions like "and", "or", "with".
- Do NOT extract abstract words like "together", "scene", "image".
- Do NOT duplicate.
- ALWAYS make count explicit: "a horse" -> "one horse".
- Keep each phrase 2-6 words.
- Output ONLY a JSON list.
\end{verbatim}

\noindent Representative outputs:
\begin{itemize}
\item ``a horse and a giraffe'' $\rightarrow$ [``one horse'', ``one giraffe''].
\item ``three red apples on a wooden table'' $\rightarrow$ [``three red apples'', ``one wooden table'', ``apples on table''].
\end{itemize}

\paragraph{Image inspection prompt.} \noindent

\noindent Input: decoded intermediate image.

\begin{verbatim}
List ALL objects and creatures visible in this image.
For EACH item, output ONE line in this exact format:
COUNT | TYPE | ATTRIBUTES

Examples:
2 | giraffe | brown and white spotted
1 | tree | green, tall
3 | apple | red, on a table
1 | hybrid creature | mix of horse and giraffe,
  spotted pattern on horse's body

Rules:
- COUNT must be a number (1, 2, 3, ...).
- TYPE must be a specific noun
  (dog, cat, tree, car, NOT "animal" or "object").
- If a creature looks abnormal, mixed, or hybrid,
  TYPE should be "hybrid creature" and describe it in ATTRIBUTES.
- Be HONEST about what you see. Do NOT guess or assume.
- If you are unsure what species an animal is,
  say "unknown animal" and describe it.
\end{verbatim}

\noindent Representative output:
\begin{verbatim}
1 | horse | brown
1 | giraffe | yellow, brown spotted
1 | tree | green
\end{verbatim}

\paragraph{Relation check prompt.} \noindent

\noindent Input: decoded intermediate image and relation phrases.

\begin{verbatim}
Look at this image carefully.
Does the image show: "<phrase>"?
Answer whether the relation is satisfied (yes or no).
\end{verbatim}

\noindent Representative example:
\begin{itemize}
\item phrase: ``cat on chair''; output: ``yes''.
\end{itemize}

Missing or incorrect required contents are mapped to $\mathcal{D}_{pos}$,
while extra unrelated contents found in image inspection are mapped to
$\mathcal{D}_{neg}$.

\setcounter{figure}{0}
\setcounter{table}{0}
\setcounter{equation}{0}
\setcounter{algorithm}{0}

\renewcommand{\thefigure}{S\arabic{figure}}
\renewcommand{\thetable}{S\arabic{table}}
\renewcommand{\theequation}{S\arabic{equation}}
\renewcommand{\thealgorithm}{S\arabic{algorithm}}

\end{document}